\documentclass[10pt,twocolumn,letterpaper]{article}
 
\usepackage[pagenumbers]{cvpr} 
 
\newcommand{\pmstd}[1]{\ensuremath{_{\pm #1}}}
\usepackage{listings}
\usepackage{xcolor}
\usepackage{listings}
\usepackage{tcolorbox}
\tcbuselibrary{breakable}
 
\usepackage{booktabs}
\usepackage{colortbl}  
\usepackage{multirow}
\usepackage{amssymb}
\usepackage{enumitem}
\usepackage{tikz}
\usepackage{pgfplots}
\pgfplotsset{compat=1.18}
\definecolor{cvprblue}{rgb}{0.21,0.49,0.74}
\usepackage[pagebackref,breaklinks,colorlinks,allcolors=cvprblue]{hyperref}
 \usepackage{fontawesome5} 
\newcommand{\hflogo}{\raisebox{-0.15em}{\includegraphics[height=1em]{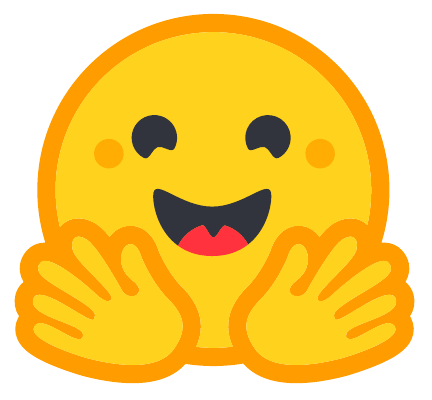}}}
\newcommand{\ghlogo}{\raisebox{-0.15em}{\includegraphics[height=1em]{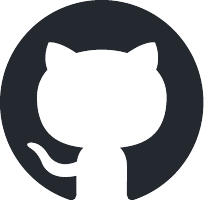}}}
\def\confName{CVPR}
\def\confYear{2027}

\usepackage{pifont}

\providecommand{\yes}{\ensuremath{\checkmark}}
\providecommand{\no}{\ensuremath{\times}}
\newcommand{\partialmark}{\ding{119}}

\newcommand{\mozaic}{\textsc{Mozaic}}
\newcommand{\mozaicsmall}{\textsc{Mozaic-Small}}
 
\title{\raisebox{-0.25em}{\includegraphics[height=1.3em]{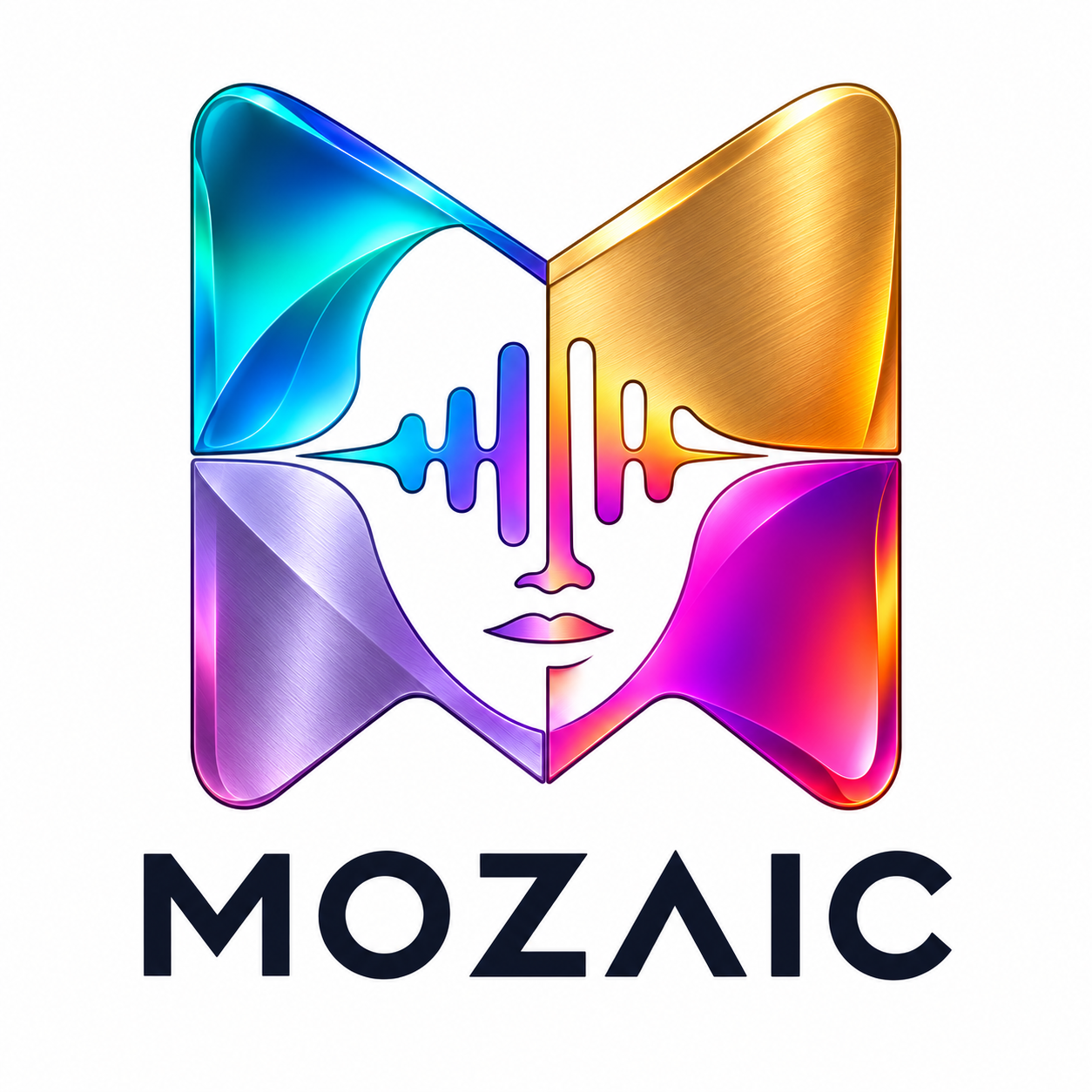}}\hspace{0.3em}\mozaic{}: An Audio-Video Deepfake Benchmark\\for Telling \emph{What} Is Fake}
 
\author{%
Shaina Raza\textsuperscript{1*}
\quad Mohamed Hafez\textsuperscript{1}
\quad Ahmed Y. Radwan\textsuperscript{1}
\quad Khalid Elgazzar\textsuperscript{2}
\quad Kathryn Hume\textsuperscript{1}\\[0.4em]
{\normalsize\textsuperscript{1}Vector Institute, Canada
\qquad \textsuperscript{2}Ontario Tech University, Canada}\\[0.3em]
{\small\ttfamily \{shaina.raza, mohamed.hafez, ahmed.radwan, kathryn.hume\}@vectorinstitute.ai}\\
{\small\ttfamily khalid.elgazzar@ontariotechu.ca}\\[0.3em]
{\small\textsuperscript{*}Corresponding author:
\href{mailto:shaina.raza@vectorinstitute.ai}{\texttt{shaina.raza@vectorinstitute.ai}}}%
}

\begin{document}
\makeatletter
\twocolumn[{\@maketitle
  \iftoggle{cvprfinal}{
    \vspace{-2.0em}\begin{center}\small
    \href{https://huggingface.co/datasets/vector-institute/Mozaic}{\hflogo\,Dataset}\qquad
    \href{https://github.com/VectorInstitute/mozaic}{\ghlogo\,Code}\qquad
    \href{https://vectorinstitute.github.io/mozaic/}{\raisebox{-0.2em}{\includegraphics[height=1em]{sec/figures/LOGO/MOZIAC-LOGO.png}}\,Project page}
    \end{center}\vspace{0.6em}}{}%
}]
\makeatother

\begin{abstract}
Audio-video (AV) deepfakes can fake the face, the voice, or both. Most benchmarks only ask whether a clip is real or fake, so they cannot show whether a detector relies on the picture, the voice, or encoding traces. We present \mozaic{}, a paired benchmark with a 4-way label in which every source clip is released in all four combinations of real or fake video and real or fake audio so each modality contribution can be measured directly. Fakes cover inner-face swapping, whole-frame video diffusion and voice conversion. \mozaic{} has 137,745 clips from 15,014 source pairs and four tasks: T1 AV authenticity, T2 manipulation type, T3 identity verification and T4 explanation. We evaluate 12 AV multimodal LLMs (MLLMs), train four detectors and fine-tune MLLMs with LoRA. Overall, the best MLLM reaches only 35.6\% on the four-way T1 and catches far more video-only fakes than voice-only fakes (43.6\% vs 7.5\%). Trained detectors reach 98.4\% on T1, and LoRA fine-tuning raises explanation scores (OmniVinci 4.17 to 7.40, Qwen2.5-Omni 4.42 to 6.98), but the best T1 detector falls to 78.3\% when the face-swap generator is held out of training, so generalization and explanation remain open. We release the dataset, the compact \mozaicsmall{} split, the code and the prompts.
\end{abstract}

\IfFileExists{sec/figures/radar_final.pdf}{%
\begin{figure}[t]
  \centering
  \includegraphics[width=\linewidth]{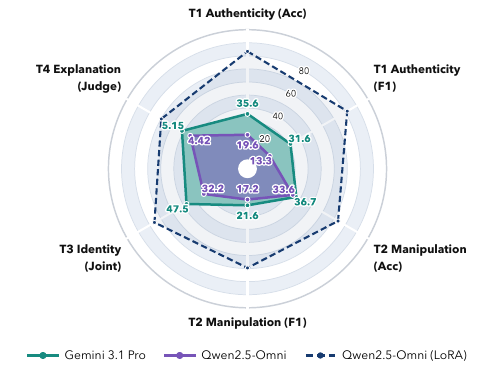}
  \vspace{-1.2em}
  \caption{\textbf{MLLMs are near chance at telling what is fake; LoRA helps but leaves a gap.} Gemini 3.1 Pro and Qwen2.5-Omni inference, and Qwen2.5-Omni after LoRA on \mozaicsmall{} (in distribution; \cref{tab:zeroshot,tab:trained_t4}).}
  \label{fig:radar}
  \vspace{-1em}
\end{figure}}{}

\section{Introduction}
\label{sec:intro}

\begin{figure*}[t]
  \centering
  \includegraphics[width=0.9\linewidth]{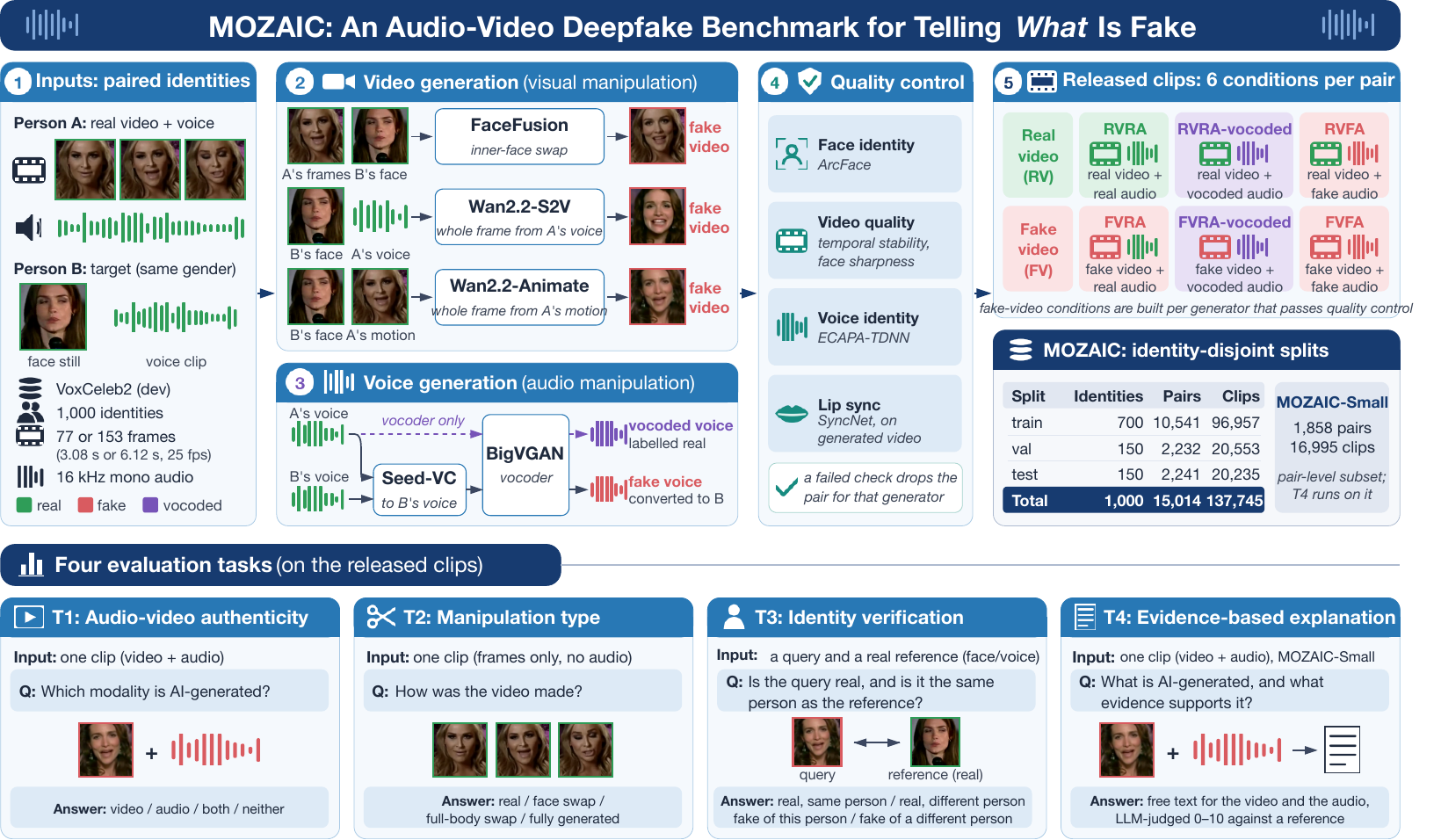}
  \vspace{-0.5em}
  \caption{\textbf{\mozaic{}.} \textit{Top:} a real clip of person A is paired with a same-gender person B; three video generators (FaceFusion, Wan2.2-S2V, Wan2.2-Animate) and one voice converter (Seed-VC) create the fakes, pairs that fail quality control are dropped, and each pair is released in six conditions (real or fake video with real, vocoded or fake audio). \textit{Bottom:} the four tasks: which modality is fake (T1), how the video was made (T2), whether a query clip is real and shows the reference person (T3), and a written explanation (T4).}
  \label{fig:teaser}
\end{figure*}

Early deepfakes manipulated the face, but recent generators go further: Wan2.2~\cite{wan2025} synthesizes entire frames, subject and background, and voice converters such as Seed-VC~\cite{seedvc} change who is speaking while keeping what is said and when. An AV deepfake can therefore have a fake video track, a fake audio track, or both~\cite{raza2026deepfakes}.

Detecting such clips calls for more than a real-or-fake decision: a detector should state which modality was manipulated, check whether two clips show the same person, and point to its evidence. MLLMs, which take audio and video together, could do all three, yet no existing benchmark tests the three together (\cref{sec:related}, \cref{tab:compare}), and real-versufake scores cannot tell whether a model relies on the video, the voice or encoding traces.

We introduce \textbf{\mozaic{}} (Modality-paired benchmark for analyZing Audio-video Integrity and Content), an AV deepfake benchmark built from 1,000 VoxCeleb2~\cite{chung2018voxceleb2} identities (\cref{fig:teaser}). Each of its 15,014 source pairs (a real clip of person A and a same-gender target B) appears in all four combinations of real or fake video and audio (a 4-way label), which isolates the effect of each modality; a vocoded real-voice control separates vocoder traces from speaker identity. \mozaic{} has 137,745 clips and four tasks: AV authenticity (T1), manipulation type (T2), identity verification (T3) and evidence-based explanation (T4, on the compact \mozaicsmall{} split). 

We evaluate 12 MLLMs, and train four detectors and LoRA-tune MLLMs under one protocol (\cref{fig:radar}). MLLMs stay close to chance at telling which modality is fake and miss most fake voices. Trained detectors nearly solve T1 and T2 in distribution but lose accuracy on an unseen face-swap generator, and explanations stay far from the ceiling after fine-tuning. Telling \emph{what} is fake, not only \emph{whether}, is still open. 

Our contributions are: (1) \mozaic{} and \mozaicsmall{}, paired AV deepfake datasets with identity-disjoint splits, released for detection research only (\cref{sec:dataset,sec:ethics}); (2) four tasks, from close-ended detection to open-ended explanation (\cref{sec:tasks}); (3) a comprehensive from evaluation of state-of-the-art  MLLMs to training four detectors and LoRA-tuning MLLMs (\cref{sec:results}); and \cref{sec:setup} the full evaluation suite for reproduction.

\section{Related Work}
\label{sec:related}

\textbf{Audio-video deepfake datasets.} FakeAVCeleb~\cite{khalid2021fakeavceleb} introduced labels RVRA, RVFA, FVRA and FVFA (real/fake video $\times$ real/fake audio), while LAV-DF~\cite{cai2022lavdf}, AV-Deepfake1M~\cite{cai2024avdeepfake1m} and AV-Deepfake1M++~\cite{avdeepfake1mpp} (about two million clips) target localized, content-driven edits. In FakeAVCeleb, LAV-DF and AV-Deepfake1M, fake audio is text-to-speech, changing speech timing and lip movements; \mozaic{} converts voices while preserving words, timing and frames, with a vocoded real-voice control. DeepSpeak~\cite{barrington2026deepspeak} adds webcam face swaps, lip-sync, avatars and voice cloning; PolyGlotFake~\cite{polyglotfake}, MAVODD~\cite{mavosdd} and BabelFake~\cite{babelfake} broaden language coverage; DeepfakeBench-MM~\cite{deepfakebenchmm} standardizes evaluation. AVFakeBench~\cite{avfakebench} tests MLLMs from binary judgment to explanation, while AVID~\cite{avid} probes AV inconsistencies. However, none jointly provides all four real/fake combinations per clip and reference-based identity checks, a gap addressed by \mozaic{}.

\textbf{Video-only deepfake datasets.} ViF-Bench~\cite{skyra} evaluates detection and visual artifact reasoning; DF26~\cite{df26} tests detector robustness across recent video generators (Wan~\cite{wan2025}, HunyuanVideo~\cite{kong2024hunyuanvideo}, Kling~\cite{klingai2026}, Veo~\cite{googleveo2026}). Both test visual authenticity only, not which modality is fake or whether two clips show the same person. \mozaic{} provide both AV deepfake clips for detection and reasoning tasks.

\textbf{Detectors and MLLMs for forensics.} We train four baselines that span AV speech and perception encoders (AV-HuBERT~\cite{shi2022avhubert}, PE-AV~\cite{vyas2026pushing}), an AV deepfake detector (HAVIC~\cite{peng2026havic}) and a visual deepfake detector (DFD-HR~\cite{sun2026dfdhr}). Identity-aware detectors such as ID-Reveal~\cite{cozzolino2021idreveal} and POI-Forensics~\cite{cozzolino2023poi} compare a clip with real footage of the same person, which our T3 (discussed in \cref{sec:tasks}) turns into a benchmark task. MLLMs also serve as explaining detectors for generated video (BusterX~\cite{busterx}), for images, audio, video and talking heads (Omni-Fake~\cite{omnifake}) and for AV deepfakes~\cite{avlmmdetect}; MADBench~\cite{madbench} tests fake audio over real video. Together, these works motivate our benchmark, but none jointly evaluates audio-video authenticity, manipulation type, crosclip identity, and explanation quality under controlled generation conditions, as summarized in
\cref{tab:compare}.

\begin{table}[t]
  \centering
  \scriptsize
  \renewcommand{\arraystretch}{0.98}
  \setlength{\tabcolsep}{2.5pt}
  \caption{\textbf{Comparison with related benchmarks.} \yes/\no/\partialmark: supported/not shown/partial. Fake V/A: fake video/audio; Full: synthesis beyond the face; Open: free text; Ref.: identity check against a reference clip; Evid.: evidence evaluation.}
  \label{tab:compare}
  \resizebox{\linewidth}{!}{%
  \begin{tabular}{@{}c@{\hspace{8pt}}lcccccc}
    \toprule
    & Benchmark & Fake V & Fake A & Full & Open & Ref. & Evid. \\
    \midrule
    \multirow{2}{*}{\rotatebox[origin=c]{90}{\textit{Video}}} & ViF-Bench~\cite{skyra}
      & \yes & \no & \yes & \partialmark & \no & \partialmark \\
     & DF26~\cite{df26}
      & \yes & \no & \yes & \no & \no & \no \\
    \midrule
    \multirow{10}{*}{\rotatebox[origin=c]{90}{\textit{Audio-video}}} & FakeAVCeleb~\cite{khalid2021fakeavceleb}
      & \yes & \yes & \no & \no & \no & \no \\
     & LAV-DF~\cite{cai2022lavdf}
      & \yes & \yes & \no & \no & \no & \no \\
     & AV-Deepfake1M++~\cite{avdeepfake1mpp}
      & \yes & \yes & \no & \no & \no & \no \\
     & DeepSpeak~\cite{barrington2026deepspeak}
      & \yes & \yes & \partialmark & \no & \no & \no \\
     & MAVODD~\cite{mavosdd}
      & \yes & \yes & \yes & \no & \no & \no \\
     & BabelFake~\cite{babelfake}
      & \yes & \yes & \partialmark & \no & \no & \no \\
     & AVFakeBench~\cite{avfakebench}
      & \yes & \yes & \yes & \yes & \no & \yes \\
     & AVID~\cite{avid}
      & \no & \yes & \no & \yes & \no & \partialmark \\
     & Omni-Fake~\cite{omnifake}
      & \yes & \yes & \yes & \yes & \no & \yes \\
     & MADBench~\cite{madbench}
      & \no & \yes & \no & \no & \no & \no \\
    \midrule
    & \textbf{\mozaic{} (ours)}
      & \yes & \yes & \yes & \yes & \yes & \yes \\
    \bottomrule
  \end{tabular}%
  }
\end{table}

\section{The \mozaic{} Benchmark}
\label{sec:benchmark}
The \mozaic{} benchmark is shown in Figure \ref{fig:teaser} and discussed in this section.
\subsection{Dataset Construction}
\label{sec:dataset}

\paragraph{Source data.}
We draw real talking-head clips from 1,000 identities in the VoxCeleb2 development set~\cite{chung2018voxceleb2} and take gender from its speaker metadata. Each clip is cut from the start of a source recording at 25\,fps and 224$\times$224, VoxCeleb2 native resolution, with 16\,kHz mono audio. We use two clip lengths, 77 frames (3.08\,s) and 153 frames (6.12\,s).

\paragraph{Pair design.}
To construct the four-way labels (RVRA, RVFA, FVRA, FVFA), each pair combines a real clip of identity A with a same-gender target identity B, as shown in \cref{fig:teaser} and Supp.~\cref{tab:pair}. Real-video clips are shared across generators, while fake-video clips are generated once per retained generator. As a control, A's voice is processed through BigVGAN~\cite{bigvgan}, preserving speaker identity while introducing the vocoder artifacts present in converted speech. T1, T3 and T4 use these vocoded voices as real audio (T2 is video-only), preventing models from identifying fakes through vocoder artifacts alone. Unvocoded clips are also released.

\begin{figure}[t]
  \centering
  \IfFileExists{sec/figures/qualitative_notitle.pdf}{\includegraphics[width=0.92\linewidth,trim=0 0 244.8bp 10bp,clip]{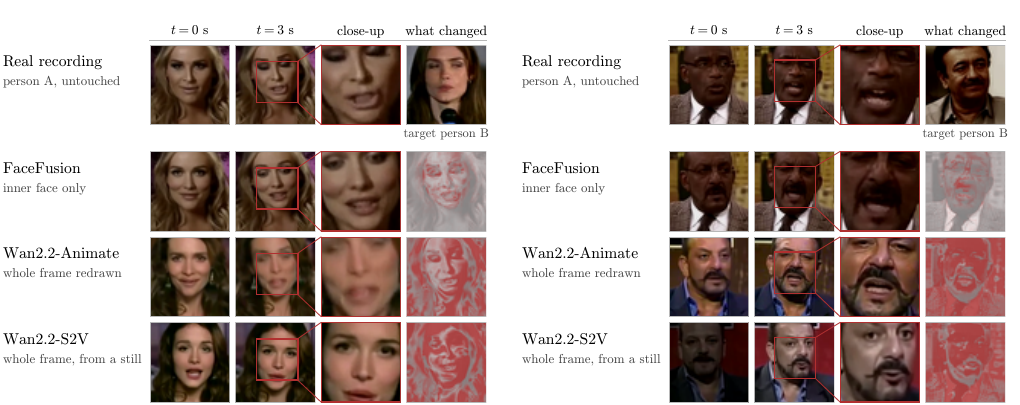}}{\includegraphics[width=0.92\linewidth,trim=0 0 244.8bp 0,clip]{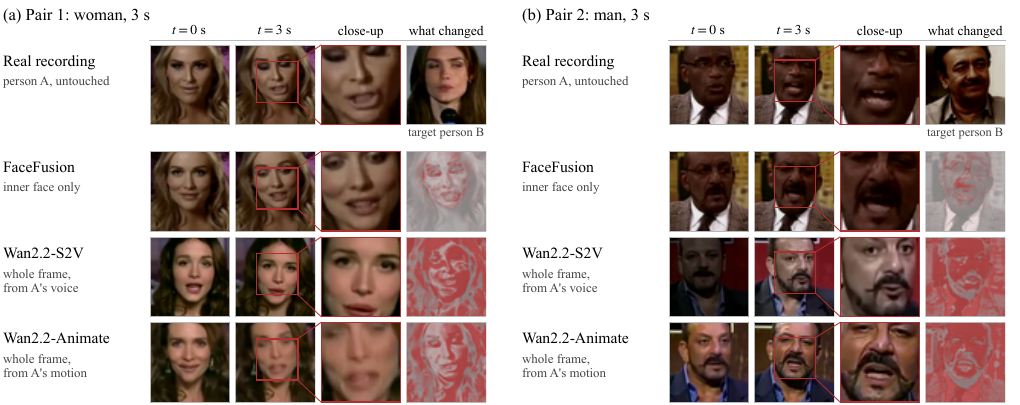}}
  \caption{\textbf{What each generator changes.} One 3\,s test pair per generator (rows), from the same real frames of A. FaceFusion replaces only the inner face; Wan2.2-Animate and Wan2.2-S2V redraw the whole frame, driven by A's motion and voice, respectively. Last column: target B (top) and changed pixels (red). Full version: Supp.~\cref{fig:qualitative_full}.}
  \label{fig:qualitative}
\end{figure}

\paragraph{Generators.}
We use three video generators and one voice converter. \textbf{FaceFusion}~\cite{facefusion} swaps the inner face, preserving the rest of the frame. \textbf{Wan2.2-S2V}~\cite{wan2025} animates an image of B using A's audio, generating the full frame. \textbf{Wan2.2-Animate}~\cite{wan2025} transfers A's motion to an image of B. \textbf{Seed-VC}~\cite{seedvc} converts A's speech into B's voice, preserving words and timing.
FaceFusion and Wan2.2-S2V support both clip lengths; Wan2.2-Animate is limited to 77 frames because longer renders caused identity drift across segments or single-pass blurring, creating potential detector shortcuts. \cref{fig:qualitative} shows one pair per generator.

\begin{figure}[t]
  \centering
  \begin{subfigure}[b]{0.44\linewidth}
    \centering
    \includegraphics[width=\linewidth]{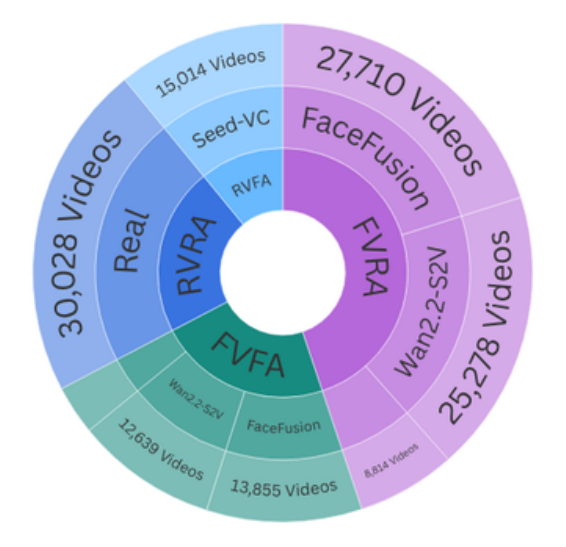}
    \caption{Clips by label and generator.}
    \label{fig:stats_clips}
  \end{subfigure}\hfill
  \begin{subfigure}[b]{0.54\linewidth}
    \centering
    \includegraphics[width=\linewidth]{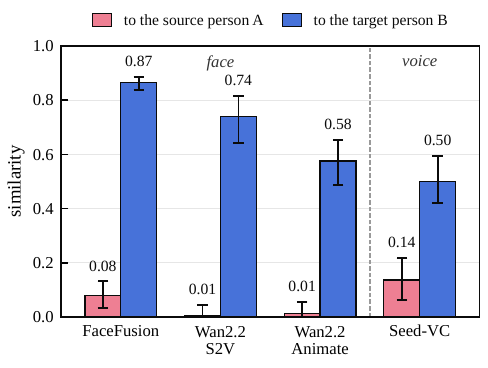}
    \caption{Identity transfer.}
    \label{fig:stats_identity}
  \end{subfigure}
  \caption{\textbf{\mozaic{} statistics.} (a)~Clips by label (inner ring) and generator (outer ring). (b)~Median similarity of each fake to the source A and the target B, face for video generators and voice for Seed-VC: every generator moves the identity to B. Question counts per task and split are in Supp.~\cref{tab:tasks}.}
  \label{fig:mozaic_stats}
  \vspace{-0.6em}
\end{figure}

\paragraph{Encoding and quality control.}
All clips share a final encoding (H.264, CRF 18; uncompressed audio), normalized brightness, color and loudness, and identical leading silence and audio fade-in. Linear probes on file, frame and audio statistics remain near chance on held-out identities (AUC $\le$ 0.57; Supp.~\cref{app:generation}), suggesting limited low-level shortcuts. ArcFace~\cite{deng2019arcface} and ECAPA-TDNN~\cite{desplanques2020ecapa} assess face and voice identity transfer to B (\cref{fig:stats_identity}); generated videos are further screened for temporal stability, sharpness and SyncNet~\cite{chung2016syncnet} lip-sync. Pairs failing any check, including failed or refused generations, are removed per generator to preserve matched conditions. This excludes 13.4\% (FaceFusion), 21.0\% (Wan2.2-S2V) and 44.9\% (Wan2.2-Animate) of 8,000 candidate pairs per applicable generator and clip length (Supp.~\cref{fig:stats,tab:rejections}).

\paragraph{Splits and statistics.}
We construct identity-disjoint training, validation and test splits with 700, 150 and 150 identities, respectively, assigning both identities in each pair to the same split. The dataset contains 15,014 source pairs and 137,745 clips, distributed as 10,541 / 2,232 / 2,241 pairs and 96,957 / 20,553 / 20,235 clips across the three splits (\cref{fig:mozaic_stats}; Supp.~\cref{tab:stats}). We also provide \mozaicsmall{}, a 16,995-clip subset sampled at the pair level, covering all identities while preserving each split's clip-length, generator and gender distributions within 0.3 percentage points of the full dataset.

\paragraph{Ethics and release.}
\label{sec:ethics}
VoxCeleb2's license permits research use. Since every fake depicts a real person, we release \mozaic{} for deepfake detection research only, with authenticity labels included in each clip's metadata and filename (Supp.~\cref{app:datasheet}).

\subsection{Tasks and Annotation}
\label{sec:tasks}
\mozaic{} defines four tasks (Supp.~\cref{tab:tasks}): T1--T3 are multiple-choice questions (MCQs) with shuffled options, and T4 is open-ended reasoning task. T1--T3 labels follow from how each clip was generated (condition, generator and pair identities); only T4 needs annotated references. The release schema is provided in Supp.~\cref{app:schema}.

\textbf{T1: Audio-video authenticity.}
Given an AV clip, the model indicates which modality, if any, was manipulated: video only (FVRA), audio only (RVFA), both (FVFA), or neither (RVRA). Audio counts as real when the speaker identity is preserved, so vocoded controls are real.

\textbf{T2: Manipulation type.}
Given video frames without audio, the model classifies the clip as \emph{real}, \emph{face swap} (FaceFusion), \emph{full-body swap} (Wan2.2-Animate, whole-frame), or \emph{fully generated} (Wan2.2-S2V); the last two separate motion-guided from audio-driven full-frame generation. Each category is defined in the prompt. Real samples come from RVRA and manipulated samples from FVRA.

\textbf{T3: Identity verification.}
Given a query clip and a real reference clip, the model decides whether the queried face or voice is real and whether it belongs to the reference person, with face and voice asked in separate questions. The four answers are \emph{real, same person}, \emph{real, different person}, \emph{fake of this person}, and \emph{fake of a different person}. 
T3 thus tests whether a model can tell a real clip of a person from a convincing fake of them.

\begin{table*}[t]
  \centering
  \setlength{\tabcolsep}{4.5pt}
  \caption{Overall results on the \mozaic{} and \mozaicsmall{} test sets, in \% except the T4 judge score. Acc: accuracy; F1: macro-F1. T3 Auth (authenticity) : is the clip real or fake; ID: do the two clips show the same person; Joint: both answers right (4-way). T4 (\mozaicsmall{} only): Judge is the 0--10 judge score, averaged over the video and audio parts; ROUGE-L is word overlap with the reference. Chance is 25\% on T1, T2 and T3 Joint and 50\% on Auth and ID; a fixed alwayreal answer scores 4.76 on Judge. \textbf{Bold}: best per column (in all tables); \underline{underlined}: second best. VideoLLaMA2.1-AV: T1 and T3 from its nopad run, T2 and T4 from the standard run.}
  \label{tab:zeroshot}
  \resizebox{\ifdim\width>\linewidth\linewidth\else\width\fi}{!}{%
  \begin{tabular}{ll | ccccccc | ccccccccc}
    \toprule
    & & \multicolumn{7}{c}{\mozaic{}} & \multicolumn{9}{c}{\mozaicsmall{}} \\
    \cmidrule(lr){3-9} \cmidrule(lr){10-18}
    & & \multicolumn{2}{c}{T1} & \multicolumn{2}{c}{T2} & \multicolumn{3}{c}{T3}
      & \multicolumn{2}{c}{T1} & \multicolumn{2}{c}{T2} & \multicolumn{3}{c}{T3} & \multicolumn{2}{c}{T4} \\
    \cmidrule(lr){3-4} \cmidrule(lr){5-6} \cmidrule(lr){7-9}
    \cmidrule(lr){10-11} \cmidrule(lr){12-13} \cmidrule(lr){14-16} \cmidrule(lr){17-18}
    Model & Size & Acc & F1 & Acc & F1 & Auth & ID & Joint & Acc & F1 & Acc & F1 & Auth & ID & Joint & Judge & ROUGE-L \\
    \midrule
    \multicolumn{18}{l}{\textit{Open models}} \\
    Qwen3-Omni           & 30B-A3B & 18.5 & 13.8 & 34.1 & 20.5 & 40.9 & 64.5 & 28.2 & 18.9 & 14.4 & 33.8 & 20.2 & 41.2 & 64.1 & 28.3 & 4.22 & 25.3 \\
    Qwen2.5-Omni         & 7B     & 19.6 & 13.3 & 33.6 & 17.2 & 57.2 & 56.8 & 32.2 & 19.5 & 12.9 & 32.3 & 16.3 & 56.8 & 57.0 & 32.1 & 4.42 & 25.1 \\
    MiniCPM-o 4.5        & 9B      & 24.0 & 18.0 & 33.5 & 15.3 & 58.3 & 56.2 & 32.3 & 23.7 & 18.0 & 33.0 & 14.9 & 58.8 & 57.0 & 32.8 & 4.39 & 22.5 \\
    Nemotron 3 Nano Omni & 30B-A3B & 18.7 & 16.3 & 33.9 & 15.2 & 57.2 & 55.5 & 32.9 & 20.2 & 17.8 & 33.6 & 14.8 & 57.0 & 55.2 & 32.6 & 4.50 & \cellcolor{cvprblue!8}\underline{26.4} \\
    Gemma 4              & 12B     & 18.2 & 11.2 & 33.3 & 12.7 & 46.2 & 60.0 & 29.6 & 18.3 & 11.4 & 33.3 & 12.9 & 46.1 & 59.9 & 30.1 & 4.12 & 24.9 \\
    VideoLLaMA2.1-AV     & 7B      & 30.9 & 18.3 & 29.6 & 11.8 & 57.5 & 51.0 & 29.2 & 31.6 & 19.2 & 29.8 & 11.8 & 57.2 & 51.8 & 29.2 & 4.41 & 25.8 \\
    OmniVinci            & 9B      & 16.6 &  8.7 & 30.0 & 11.8 & \cellcolor{cvprblue!8}\underline{60.1} & 50.5 & 30.0 & 16.6 &  8.6 & 30.0 & 11.9 & \cellcolor{cvprblue!8}\underline{60.0} & 50.8 & 30.0 & 4.17 & 23.5 \\
    Uni-MoE-2.0-Omni     & 33B     & 29.8 & 22.1 & 31.2 & 18.5 & 48.4 & 53.5 & 26.1 & 29.4 & 21.3 & 30.5 & 17.6 & 48.2 & 54.8 & 26.4 & 4.33 & 23.5 \\
    Ola                  & 7B      & \cellcolor{cvprblue!8}\underline{32.5} & 13.7 & 32.7 & 18.7 & 39.9 & 53.5 & 22.4 & \cellcolor{cvprblue!8}\underline{32.7} & 13.9 & 32.3 & 18.3 & 39.9 & 53.1 & 22.2 & 4.02 & 23.2 \\
    VITA-1.5             & 7B      & 17.3 &  9.4 & 32.0 & 19.6 & 40.5 & 53.8 & 22.5 & 17.4 &  9.6 & 32.2 & 19.7 & 40.6 & 52.8 & 22.3 & 4.52 & 24.4 \\
    \midrule
    \multicolumn{18}{l}{\textit{Proprietary models}} \\
    Gemini 3.8 Flash     & --      & 31.6 & \cellcolor{cvprblue!8}\underline{24.5} & \cellcolor{cvprblue!20}\textbf{41.0} & \cellcolor{cvprblue!20}\textbf{25.3} & 50.1 & \cellcolor{cvprblue!8}\underline{70.2} & \cellcolor{cvprblue!8}\underline{40.9} & 32.0 & \cellcolor{cvprblue!8}\underline{24.8} & \cellcolor{cvprblue!20}\textbf{41.4} & \cellcolor{cvprblue!20}\textbf{25.7} & 50.1 & \cellcolor{cvprblue!8}\underline{69.7} & \cellcolor{cvprblue!8}\underline{40.5} & \cellcolor{cvprblue!8}\underline{5.09} & 24.2 \\
    Gemini 3.1 Pro       & --      & \cellcolor{cvprblue!20}\textbf{35.6} & \cellcolor{cvprblue!20}\textbf{31.6} & \cellcolor{cvprblue!8}\underline{36.7} & \cellcolor{cvprblue!8}\underline{21.6} & \cellcolor{cvprblue!20}\textbf{62.7} & \cellcolor{cvprblue!20}\textbf{71.3} & \cellcolor{cvprblue!20}\textbf{47.5} & \cellcolor{cvprblue!20}\textbf{35.8} & \cellcolor{cvprblue!20}\textbf{31.7} & \cellcolor{cvprblue!8}\underline{37.0} & \cellcolor{cvprblue!8}\underline{21.6} & \cellcolor{cvprblue!20}\textbf{62.7} & \cellcolor{cvprblue!20}\textbf{71.3} & \cellcolor{cvprblue!20}\textbf{47.1} & \cellcolor{cvprblue!20}\textbf{5.15} & \cellcolor{cvprblue!20}\textbf{27.8} \\
    \bottomrule
  \end{tabular}}
\end{table*}

\textbf{T4: Evidence-based explanation.}
Given an AV clip from \mozaicsmall{}, the model explains video and audio separately, identifying whether each is manipulated and its scope (\emph{inner face} or \emph{whole frame} for video; \emph{voice} for audio). Reference explanations are grounded in clip metadata. Gemini~3.6 Flash drafts each explanation using eight frames and the audio, while Claude Sonnet~5.5 verifies it against the metadata and standardizes the wording (Supp.~\cref{app:t4-annotation}). Three authors reviewed a stratified sample of 400 references, accepting 99.2\% (Gwet's AC1~\cite{gwet2008computing}: 0.985 for video and 0.983 for audio).

An LLM judge (GPT-5.4 mini) scores explanations against their references on a 0--10 rubric (Supp.~\cref{app:prompts,tab:t4}). Claude and the judge belong to model families excluded from evaluation. A manual review of 400 judged explanations across several models found scores consistent with the rubric.


\section{Experimental Setup}
\label{sec:setup}
\mozaic{} experimental design is given in this section.

\subsection{Models and Data}
T1--T3 are evaluated on the test splits of both \mozaic{} and \mozaicsmall{}, while T4 uses only \mozaicsmall{}.
We evaluate 12 MLLMs zero-shot: ten open-source models (Qwen3-Omni~\cite{xu2025qwen3omni}, Qwen2.5-Omni~\cite{qwen25omni}, MiniCPM-o 4.5~\cite{cui2026minicpm}, Nemotron 3 Nano Omni~\cite{deshmukh2026nemotron}, Gemma 4~\cite{gemma4}, VideoLLaMA2.1-AV~\cite{cheng2024videollama2}, OmniVinci~\cite{ye2025omnivinci}, Uni-MoE-2.0-Omni~\cite{li2025uni}, Ola~\cite{liu2025ola} and VITA-1.5~\cite{fu2025vita}) and two proprietary models (Gemini 3.8 Flash and Gemini 3.1 Pro~\cite{team2023gemini}). Ablations use five models selected for cost and coverage: Gemini 3.8 Flash, Qwen3-Omni, Nemotron 3 Nano Omni, MiniCPM-o 4.5 and Ola.

Trained baselines use the \mozaicsmall{} training split and are evaluated on both datasets. AV-HuBERT~\cite{shi2022avhubert}, PE-AV~\cite{vyas2026pushing} and HAVIC~\cite{peng2026havic} use classification heads with full encoder fine-tuning (Full FT) or LayerNorm-only adaptation (LN). T1 uses audio and video; T2 is video-only and additionally includes DFD-HR~\cite{sun2026dfdhr}, trained with adapters. For T3, a shared encoder embeds the test and reference clips, and a classification head predicts one of four answers from their concatenated embeddings. Qwen2.5-Omni is LoRA-tuned on all four tasks. For T4, six  open MLLMs (Qwen3-Omni, MiniCPM-o 4.5, Nemotron 3 Nano Omni, Gemma 4, OmniVinci and Ola) are LoRA-tuned on the \mozaicsmall{} training split using the same settings.

Hyperparameters are provided in Supp.~\cref{app:compute,tab:training}, with learning-rate and epoch sweeps in Supp.~\cref{tab:sweeps}. To assess generalization, we retrain each baseline while excluding one video generator and evaluate on its fake clips alongside all real clips (\cref{fig:seen_unseen}).

\subsection{Input Protocol}
Models run through their native multimodal interfaces and official preprocessing. Every model receives 32 frames per clip: models with native video input sample them with their own preprocessing, and the others receive 32 uniformly sampled frames; audio is provided for AV tasks and omitted for T2. T3 receives both clips, as frames for face questions and audio for voice questions (Supp.~\cref{app:prompts}). Per-model input settings and actual frame counts are reported in Supp.~\cref{app:models}.

\subsection{Decoding}
Generative models use task-specific prompts (Supp.~\cref{app:prompts}) with temperature 0. For T1--T3, models return a JSON reply naming one option, with decoding restricted to the valid answers for MiniCPM-o 4.5, Uni-MoE-2.0-Omni, Ola and VITA-1.5, whose unconstrained replies often did not follow the JSON format; for T4, models generate explanations. Unparseable responses are counted as incorrect, and T4 parse rates are reported in Supp.~\cref{tab:t4}. Trained classification models output class predictions.

\subsection{Metrics}
T1 and T2 are evaluated using accuracy and macro-F1. T3 reports authenticity accuracy, identity accuracy, and joint accuracy on the 4-way answer. Trained models report accuracy and macro-F1 on T1--T2, joint accuracy on T3, and the judge score on T4 (\cref{sec:tasks}). We also report real/fake accuracy and ROUGE-L~\cite{lin2004rouge}, BLEU-4~\cite{papineni2002bleu} and CIDEr-D~\cite{vedantam2015cider} (Supp.~\cref{tab:t4}).

\begin{figure*}[t]
  \centering
  \includegraphics{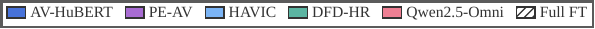}\\[3pt]
  \begin{subfigure}[b]{2.29in}\centering
    \includegraphics{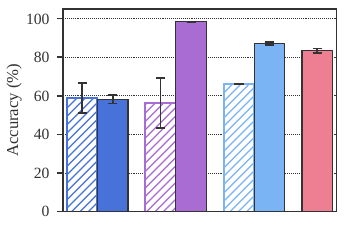}
    \caption{T1: Authenticity}
  \end{subfigure}\hfill
  \begin{subfigure}[b]{2.09in}\centering
    \includegraphics{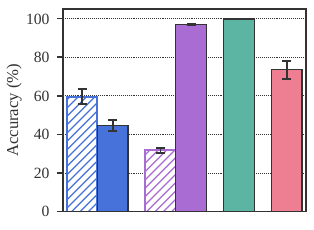}
    \caption{T2: Manipulation type}
  \end{subfigure}\hfill
  \begin{subfigure}[b]{1.77in}\centering
    \includegraphics{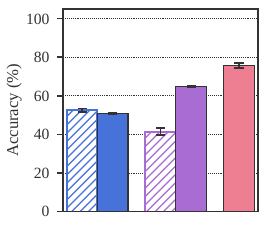}
    \caption{T3: Identity}
  \end{subfigure}
  \caption{\textbf{Seed stability of the trained models.} Mean accuracy over 5 seeds ($\pm$1 std), trained on \mozaicsmall{} and tested on \mozaic{}. Solid: parameter-efficient tuning (LN for AV-HuBERT, PE-AV and HAVIC; adapters for DFD-HR); hatched: full fine-tuning. Exact values, including F1 and T4, are in Supp.~\cref{tab:seed_stability}.}
  \label{fig:seeds}
\end{figure*}

\section{Results}
\label{sec:results}
We organize the results around four research questions: (\textbf{RQ1}) can off-the-shelf MLLMs tell what is fake; (\textbf{RQ2}) does training solve the tasks and generalize to unseen generators; (\textbf{RQ3}) what evidence do the models rely on across generators, modalities, clip lengths, and frame budgets; and (\textbf{RQ4}) can models explain and localize a manipulation.

\subsection{Evaluation of MLLMs}

\Cref{tab:zeroshot} reports the performance of all twelve MLLMs.
On \textbf{T1}, Gemini~3.1 Pro attains the highest accuracy (35.6\%) and macro-F1 (31.6\%). It identifies 43.6\% of video-only fakes but only 7.5\% of voice-only fakes, and several open models fall below the 25\% chance level because they answer ``real'' for nearly every clip (Supp.~\cref{tab:confusion_all}). Among the open models, Ola reaches the highest accuracy (32.5\%) but only 13.7\% macro-F1, because it predicts FVRA for 96\% of clips (Supp.~\cref{fig:confusion}).
On \textbf{T2}, Gemini~3.8 Flash performs best (41.0\%), whereas every open model remains below 35\%. On \textbf{T3}, Gemini~3.1 Pro achieves 62.7\% authenticity accuracy and 71.3\% identity accuracy, but only 47.5\% joint accuracy. Results on \mozaicsmall{} follow the same trends.
On \textbf{T4}, Gemini~3.1 Pro again obtains the highest explanation score (5.15/10) and ROUGE-L (27.8). \textbf{Takeaway (RQ1):} Although the Gemini models perform best on every task, no model reaches 50\% on the four-way T1 and T2 answers or on T3's joint answer, or exceeds 5.2 out of 10 on T4.

 \begin{table}[t]
  \centering
  \scriptsize
\caption{\textbf{Supervised fine-tuning on \mozaicsmall{}.} T4 judge scores (0--10) before and after LoRA SFT; $\Delta$: improvement. Trained T1--T3 results appear in Supp.~\cref{tab:seed_stability}.}
  \label{tab:trained_t4}
  \setlength{\tabcolsep}{4pt}
  \resizebox{\linewidth}{!}{%
    \begin{tabular}{lccc}
    \toprule
    Model & Zero-shot & LoRA SFT & $\Delta$ \\
    \midrule
    Qwen3-Omni           & 4.22 & 6.23 & +2.01 \\
    Qwen2.5-Omni         & 4.42 & 6.98 & +2.56 \\
    MiniCPM-o 4.5        & 4.39 & 6.03 & +1.64 \\
    Nemotron 3 Nano Omni & \textbf{4.50} & 5.18 & +0.68 \\
    Gemma 4              & 4.12 & 5.23 & +1.11 \\
    OmniVinci            & 4.17 & \textbf{7.40} & \textbf{+3.23} \\
    Ola                  & 4.02 & 5.40 & +1.38 \\
    \bottomrule
  \end{tabular}
  }
\end{table}

\subsection{Supervised Baseline Evaluation}
\begin{figure}[t]
  \centering
  \includegraphics[width=\linewidth]{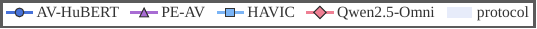}\\[2pt]
  \begin{subfigure}[b]{0.49\linewidth}
    \centering
    \includegraphics[width=\linewidth]{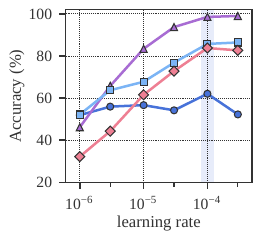}
    \caption{Learning rate.}
    \label{fig:sensitivity_lr}
  \end{subfigure}\hfill
  \begin{subfigure}[b]{0.49\linewidth}
    \centering
    \includegraphics[width=\linewidth]{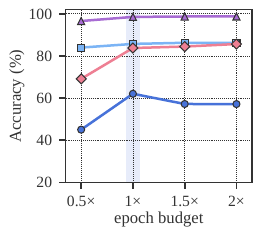}
    \caption{Epoch budget.}
    \label{fig:sensitivity_epochs}
  \end{subfigure}
  \caption{\textbf{Sensitivity to learning rate and training length}, T1 test accuracy. (a)~Learning rate (log scale); (b)~number of epochs relative to the protocol, with all other settings fixed. AV-HuBERT, PE-AV and HAVIC use LN tuning; Qwen2.5-Omni uses LoRA. Shaded: protocol value. Full results for all tasks (T1--T4) and settings, including Full FT, are in Supp.~\cref{tab:sweeps}.}
  \label{fig:sensitivity}
\end{figure}

\cref{fig:seeds} and Supp.~\cref{tab:seed_stability} report the trained models on T1--T3 (means over five seeds), \cref{tab:trained_t4} reports T4, and \cref{fig:sensitivity} shows their sensitivity to learning rate and epoch budget.
\textbf{In distribution.} LoRA SFT raises the T4 score of every tuned MLLM (\cref{tab:trained_t4}): OmniVinci gains the most (4.17 to 7.40), and Nemotron 3 Nano Omni, the strongest of these MLLMs off-the-shelf, the least (+0.68). Qwen2.5-Omni improves from 4.42 to 6.98, alongside gains from 19.6\% to 83.4\% on T1, 33.6\% to 73.6\% on T2 and 32.2\% to 75.8\% on T3 (joint). T3 stays the hardest task after training: LN-tuned PE-AV reaches 64.8\% joint accuracy and AV-HuBERT about 51\% (Supp.~\cref{tab:seed_stability}). 

Across five seeds, full fine-tuning shows greater T1 variability, especially for PE-AV and AV-HuBERT (\cref{fig:seeds}). With LN tuning, PE-AV reaches 98.4\% on T1 and 96.8\% on T2 with little seed variation, suggesting effective feature transfer. AV-HuBERT remains less effective, possibly due to its speech-focused pretraining. HAVIC reaches 86.9\% on T1 with LN tuning, while DFD-HR reaches 99.9\% on T2 with adapters. Even after LoRA tuning, Qwen2.5-Omni trails these specialist detectors (83.4\% vs.\ 98.4\% on T1, 73.6\% vs.\ 99.9\% on T2). Full fine-tuning is unstable at the protocol learning rate: lowering it to $10^{-5}$ lifts PE-AV from 56.4\% (5-seed mean, Supp.~\cref{tab:seed_stability}) to 99.8\% on T1 (sweeps in \cref{fig:sensitivity} and Supp.~\cref{tab:sweeps}).

\begin{figure}[h]
  \centering
  \includegraphics[width=\linewidth]{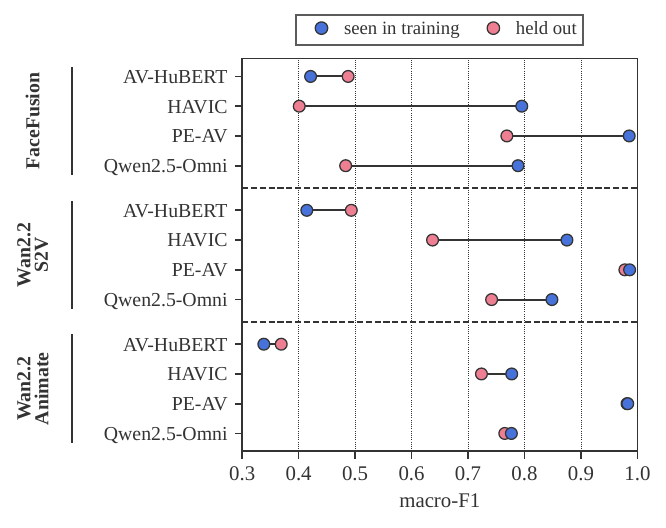}
  \caption{\textbf{Generalization to unseen generators.}
  Macro-F1 (4-class) of each detector when a generator is seen in training (blue) and when it is held out from training (pink); each group holds out one generator and evaluates on its clips. The line length shows the drop. Detailed results for all metrics and configurations are in Supp.~\cref{tab:seen_unseen_full}.}
  \label{fig:seen_unseen}
\end{figure}

\textbf{Unseen generators.} This in-distribution accuracy does not carry over to an unseen face-swap generator (\cref{fig:seen_unseen}; Supp.~\cref{tab:seen_unseen_full}). When FaceFusion is held out of training, T1 macro-F1 on its clips falls from 98.6 to 76.9 for PE-AV, from 79.5 to 40.1 for HAVIC, and from 78.9 to 48.3 for Qwen2.5-Omni, whereas holding out either whole-frame generator costs PE-AV less than one point. \textbf{Takeaway (RQ2):} In-distribution accuracy overstates robustness: the detectors lose 22--39 F1 points on an unseen inner-face swap, which suggests that they rely on generator-specific cues.


 \begin{table}[h]
  \centering
  \small
  \caption{\textbf{Modality ablation.} T1 balanced real/fake accuracy (\%; chance 50) on each modality, \mozaicsmall{} test set, with both inputs (A+V), frames only (V) and sound only (A). Bold: best per column, where above chance. Full results: Supp.~\cref{tab:audioablation}.}
  \label{tab:modality}
  \setlength{\tabcolsep}{4pt}
  \resizebox{\ifdim\width>\linewidth\linewidth\else\width\fi}{!}{%
  \begin{tabular}{lcccccc}
    \toprule
    & \multicolumn{3}{c}{Video accuracy} & \multicolumn{3}{c}{Audio accuracy} \\
    \cmidrule(lr){2-4} \cmidrule(lr){5-7}
    Model & A+V & V & A & A+V & V & A \\
    \midrule
    Gemini 3.8 Flash     & \textbf{64.0} & \textbf{59.4} & 50.2 & \textbf{52.1} & 50.1 & 51.8 \\
    Nemotron 3 Nano Omni & 51.8 & 53.1 & 49.8 & 48.9 & 50.0 & 49.0 \\
    Ola                  & 49.9 & 50.0 & 50.5 & 49.9 & 50.0 & 51.3 \\
    MiniCPM-o 4.5        & 52.8 & 52.2 & 49.7 & 50.4 & 50.0 & 45.9 \\
    Qwen3-Omni           & 51.5 & 52.5 & 50.0 & 51.8 & 50.0 & \textbf{58.6} \\
    \bottomrule
  \end{tabular}}
\end{table}

\subsection{Modality Use}
To measure how much each modality contributes, we repeat T1 on \mozaicsmall{} with video-only and audio-only inputs for the five-model subset, stating the missing modality in the prompt (\cref{tab:modality}; Supp.~\cref{tab:audioablation}). Removing the audio changes four-class accuracy by at most 3.5 percentage points and even improves it for three of the five models. Qwen3-Omni benefits most from audio-only input: its balanced audio accuracy rises from 51.8\% to 58.6\%, whereas every other audio-accuracy cell remains near chance (45.9--52.1\%). In contrast, Gemini~3.8 Flash achieves 64.0\% balanced video accuracy with both modalities and 59.4\% with video alone, but drops to 50.2\% once the video is removed. \textbf{Takeaway (RQ3):} Of the five models, only Gemini~3.8 Flash makes clear use of visual evidence, and none reliably detects audio manipulation when video is present.

 \begin{figure}[h]
  \centering
  \includegraphics[width=\linewidth]{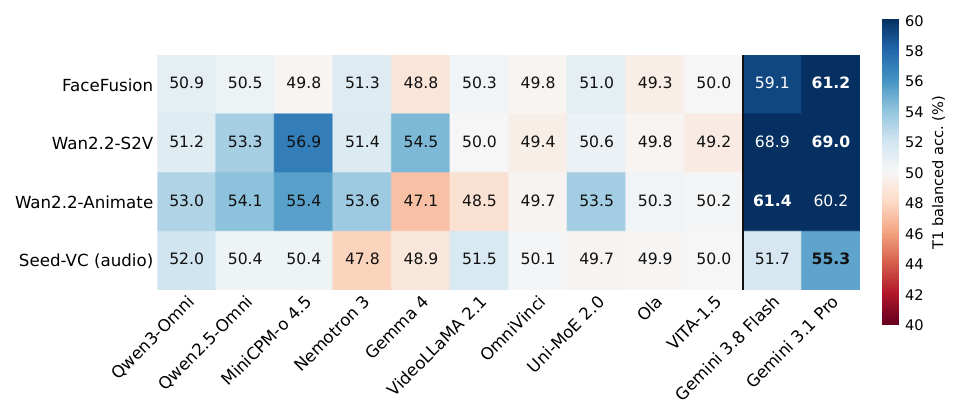}
  \caption{\textbf{T1 balanced accuracy (\%) per generator} for every off-the-shelf model on the \mozaic{} test set. Each cell averages the fake-part recall for that generator with the real-part recall of the same modality (video; audio for Seed-VC), so a constant answer scores 50\% (chance). Red is below chance, blue above, saturating at $\pm$10 points; bold marks the best model per generator. Nemotron 3, VideoLLaMA 2.1 and Uni-MoE 2.0 abbreviate Nemotron 3 Nano Omni, VideoLLaMA2.1-AV (nopad run, as in \cref{tab:zeroshot}) and Uni-MoE-2.0-Omni.}
  \label{fig:pergen}
\end{figure}
\begin{figure*}[t]
  \centering
  \IfFileExists{sec/figures/t4_qualitative_v4.pdf}{\includegraphics[width=\textwidth]{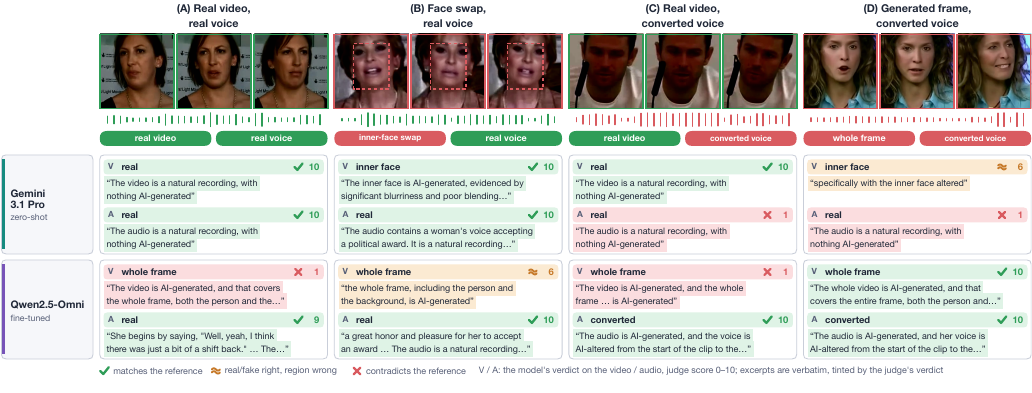}}{\includegraphics[width=0.8\textwidth]{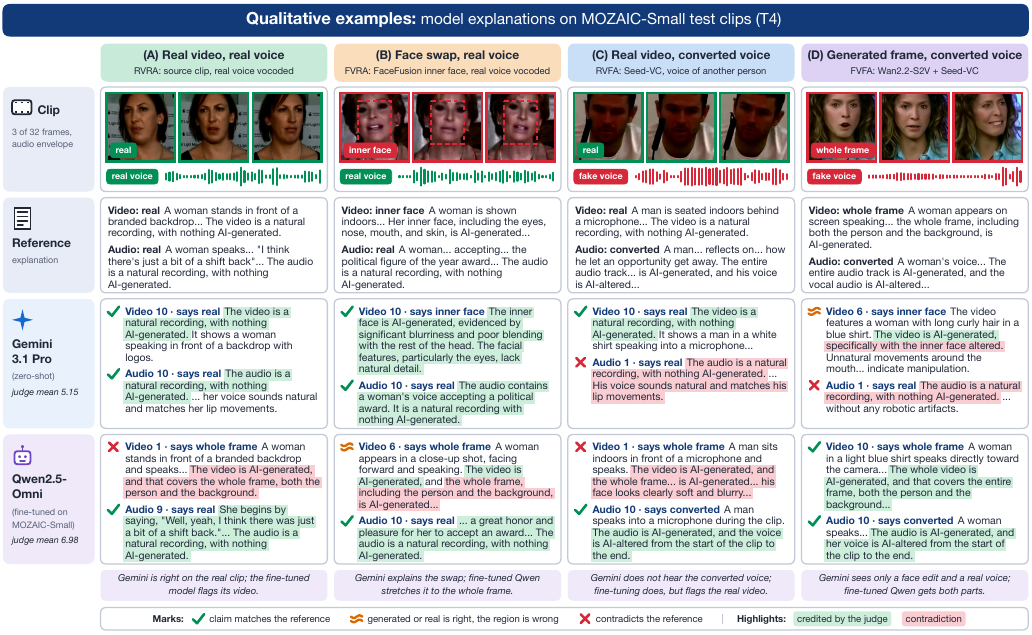}}
  \caption{\textbf{T4 on four \mozaicsmall{} test clips:} real, face swap with real voice, real video with converted voice, and fully generated. Borders, envelopes and pills: green real, red generated; dashed box: swapped face. Each chip gives what the model says about the video (V) or audio (A), a mark (check: matches the reference; wave: right verdict, wrong region; cross: contradicts it) and the judge's 0--10 score, followed by a verbatim excerpt. Full explanations: Supp.~\cref{fig:t4_full}; whole-test-set rates: Supp.~\cref{fig:t4_claims}.}
  \vspace{-0.5em}
  \label{fig:t4}
\end{figure*}

\subsection{Per-Generator Performance}

\cref{fig:pergen} reports balanced accuracy, the mean of the true-positive and true-negative rates, for each generator. The Gemini models discriminate best, reaching about 69\% on Wan2.2-S2V and 59--61\% on FaceFusion and Wan2.2-Animate, while Seed-VC remains the most difficult generator (55\% for Gemini~3.1 Pro). For the open models, 36 of 40 model--generator combinations lie within four percentage points of chance. The one open exception is MiniCPM-o~4.5 on the whole-frame generators (56.9\% on Wan2.2-S2V, 55.4\% on Wan2.2-Animate). The converted voice is the blind spot for all models: no open model exceeds 52\% on Seed-VC, and Gemini~3.8 Flash reaches only 51.7\%. \textbf{Takeaway (RQ3):} The four-class accuracy of the open models reflects prediction bias rather than genuine forgery discrimination.

\begin{figure}[t]
  \centering
  \includegraphics[width=\linewidth]{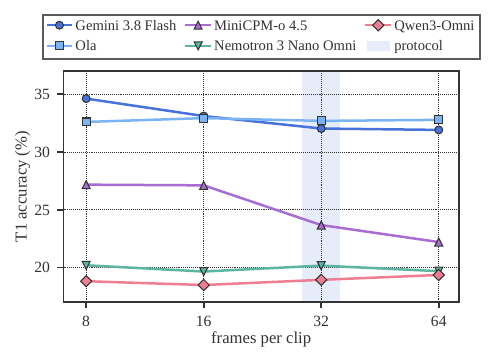}
  \caption{T1 accuracy (\%) on the \mozaicsmall{} against frames per clip (log scale). Shaded: the 32-frame protocol.}
  \label{fig:frames}
\end{figure}

\subsection{Effect of Frame Budget}
Varying the number of frames per clip from 8 to 64 for T1 on \mozaicsmall{} across five models (\cref{fig:frames}), three models change by less than two percentage points, while MiniCPM-o~4.5 drops from 27.2\% to 22.2\% and Gemini~3.8 Flash from 34.6\% to 31.9\%. \textbf{Takeaway (RQ3):} Increasing the frame budget does not improve performance, suggesting that limited frame sampling is not the main cause of poor zero-shot results.

\subsection{Effect of Clip Length}
We next examine clip duration by comparing the 3\,s and 6\,s clips of the \mozaic{} test set for all models (Supp.~\cref{tab:length}). Longer clips appear to improve T2 by 5--8 percentage points for every model, including Gemini~3.8 Flash (37.4\% to 45.5\%). However, the 6\,s clips contain no Wan2.2-Animate, one of the four T2 classes, and every model scores at most 0.3\% on Wan2.2-Animate clips (Supp.~\cref{tab:pergen_t2}), so dropping that class raises the average by construction. Once Wan2.2-Animate is also excluded from the 3\,s clips, the T2 change shrinks to between $-1.2$ and $+2.0$ points (mean $+0.1$; Supp.~\cref{tab:length_ctrl}); on T1 and T4, where no class is removed, 6\,s clips move accuracy by at most 4.3 points and the judge score by at most 0.6. \textbf{Takeaway (RQ3):} The apparent gains stem from generator composition rather than clip duration.

 \subsection{Explanation Quality}

\cref{fig:t4} shows the verdicts of Gemini~3.1 Pro and the fine-tuned Qwen2.5-Omni on four test clips, with an excerpt of each explanation, and Supp.~\cref{fig:t4_claims} summarizes the region that each model claims across the whole test set. The open MLLMs label nearly every clip as real. On the judge, every model (at most 4.52) scores below a fixed alwayreal answer (4.76), and the Gemini models fall below it on audio. Gemini~3.1 Pro names a swapped inner face in about half of all clips, including real ones, and never a generated whole frame. Qwen2.5-Omni fine-tuned on \mozaicsmall{} separates real from converted voices but labels most videos as whole-frame generated, regardless of their true type. After fine-tuning, real/fake accuracy reaches 86.8\% on audio but only 65.0\% on video. Per-part scores and text metrics for all models are given in Supp.~\cref{tab:t4}. \textbf{Takeaway (RQ4):} No model, whether untuned or fine-tuned, reliably localizes the manipulation.

\section{Conclusion}
\label{sec:conclusion}
We present \mozaic{}, an AV deepfake benchmark of 137,745 clips from 15,014 source pairs, covering all four video/audio real/fake combinations and a vocoded control. It supports four tasks: AV authenticity (T1), manipulation type (T2), identity verification (T3) and explanation (T4). State-of-the-art MLLMs struggle: the best achieves only 35.6\% on T1, fake voices are frequently missed, and open models' explanations score below an alwayreal baseline. Trained detectors reach 98.4\% in distribution but lose up to 20 percentage points on an unseen face-swap generator. These findings highlight the need for modality-specific and crosgenerator evaluation beyond aggregate accuracy.

\textbf{Limitations.}
\mozaic{} relies on VoxCeleb2 (224$\times$224, 3--6\,s clips), one voice converter and three video generators. Some compression and voice-conversion artifacts persist despite normalization. T4 references are LLM-generated (400 human-checked), while judge validation lacks a formal agreement study.

\textbf{Future Directions.}
We release \mozaic{} and its evaluation suite to support reproducible comparisons. Future work should expand source diversity, generators and clip lengths, improve artifact control, and develop detectors that generalize across generators, identify altered modalities and provide verifiable explanations.
\section*{Acknowledgments}
Resources used in preparing this research were provided, in part, by the Province of Ontario and the Government of Canada through CIFAR, as well as companies sponsoring the Vector Institute (\url{http://www.vectorinstitute.ai/\#partners}).
This research was funded by the European Union's Horizon Europe research and innovation programme under the AIXPERT project (Grant Agreement No.\ 101214389), which aims to develop an agentic, multi-layered, GenAI-powered framework for creating explainable, accountable, and transparent AI systems.

\section*{LLM Usage Statement}
During the preparation of this work, the authors used Claude Sonnet and ChatGPT to improve grammar and wording. All final text was reviewed and approved by the authors, who maintain full responsibility for the content of the manuscript.

{
    \small
    \bibliographystyle{ieeenat_fullname}
    \bibliography{references}

@inproceedings{chung2018voxceleb2,
      author       = "Chung, J.~S. and Nagrani, A. and Zisserman, A.",
      title        = "VoxCeleb2: Deep Speaker Recognition",
      booktitle    = "INTERSPEECH",
      year         = "2018",
    }

@article{khalid2021fakeavceleb,
  title={{FakeAVCeleb}: A novel audio-video multimodal deepfake dataset},
  author={Khalid, Hasam and Tariq, Shahroz and Kim, Minha and Woo, Simon S},
  journal={arXiv preprint arXiv:2108.05080},
  year={2021}
}

@misc{googleveo2026,
  author       = {{Google DeepMind}},
  title        = {Veo: AI Video Generation},
  year         = {2026},
  howpublished = {\url{https://deepmind.google/models/veo/}},
  note         = {Accessed: 2026-11-11}
}

@misc{klingai2026,
  author       = {{Kling AI}},
  title        = {Kling AI: AI Video and Image Generation Platform},
  year         = {2026},
  howpublished = {\url{https://kling.ai}},
  note         = {Accessed: 2026-11-11}
}

@article{kong2024hunyuanvideo,
  title={Hunyuanvideo: A systematic framework for large video generative models},
  author={Kong, Weijie and Tian, Qi and Zhang, Zijian and Min, Rox and Dai, Zuozhuo and Zhou, Jin and Xiong, Jiangfeng and Li, Xin and Wu, Bo and Zhang, Jianwei and others},
  journal={arXiv preprint arXiv:2412.03603},
  year={2024}
}

@inproceedings{cai2022lavdf,
  title={Do you really mean that? content driven audio-visual deepfake dataset and multimodal method for temporal forgery localization},
  author={Cai, Zhixi and Stefanov, Kalin and Dhall, Abhinav and Hayat, Munawar},
  booktitle={2022 International Conference on Digital Image Computing: Techniques and Applications (DICTA)},
  pages={1--10},
  year={2022},
  organization={IEEE}
}

@inproceedings{cai2024avdeepfake1m,
  title={{AV-Deepfake1M}: A large-scale LLM-driven audio-visual deepfake dataset},
  author={Cai, Zhixi and Ghosh, Shreya and Adatia, Aman Pankaj and Hayat, Munawar and Dhall, Abhinav and Gedeon, Tom and Stefanov, Kalin},
  booktitle={Proceedings of the 32nd ACM international conference on multimedia},
  pages={7414--7423},
  year={2024}
}

@inproceedings{avdeepfake1mpp,
  title={{AV-Deepfake1M++}: A large-scale audio-visual deepfake benchmark with real-world perturbations},
  author={Cai, Zhixi and Kuckreja, Kartik and Ghosh, Shreya and Chuchra, Akanksha and Khan, Muhammad Haris and Tariq, Usman and Gedeon, Tom and Dhall, Abhinav},
  booktitle={Proceedings of the 33rd ACM International Conference on Multimedia},
  pages={13686--13691},
  year={2025}
}

@inproceedings{barrington2026deepspeak,
  title={The {DeepSpeak} dataset},
  author={Barrington, Sarah and Bohacek, Maty and Farid, Hany},
  booktitle={Proceedings of the IEEE/CVF Conference on Computer Vision and Pattern Recognition},
  pages={1893--1902},
  year={2026}
}

@inproceedings{polyglotfake,
  title={{PolyglotFake}: A novel multilingual and multimodal deepfake dataset},
  author={Hou, Yang and Fu, Haitao and Chen, Chunkai and Li, Zida and Zhang, Haoyu and Zhao, Jianjun},
  booktitle={International conference on pattern recognition},
  pages={180--193},
  year={2024},
  organization={Springer}
}

@article{mavosdd,
  title={{MAVOS-DD}: Multilingual audio-video open-set deepfake detection benchmark},
  author={Croitoru, Florinel-Alin and Hondru, Vlad and Popescu, Marius and Ionescu, Radu Tudor and Khan, Fahad Shahbaz and Shah, Mubarak},
  journal={arXiv preprint arXiv:2505.11109},
  year={2025}
}

@article{babelfake,
  title={{BabelFake}: A Multilingual Audio-Visual DeepFake Benchmark},
  author={Segna, Carlotta and Tschesche, Joel and Rohrbach, Anna},
  journal={arXiv preprint arXiv:2610.06339},
  year={2026}
}

@inproceedings{skyra,
  title={Skyra: {AI}-generated video detection via grounded artifact reasoning},
  author={Li, Yifei and Zheng, Wenzhao and Zhang, Yanran and Sun, Runze and Zheng, Yu and Chen, Lei and Zhou, Jie and Lu, Jiwen},
  booktitle={Proceedings of the IEEE/CVF Conference on Computer Vision and Pattern Recognition},
  pages={4482--4493},
  year={2026}
}

@article{df26,
  title={DF26: We Cannot Tell Fake From Real Anymore},
  author={Shykula, Severyn and Yermakov, Andrii and Samarskyi, Ivan and Mishkin, Dmytro and Cech, Jan and Mishchuk, Anastasiia},
  journal={arXiv preprint arXiv:2609.07369},
  year={2026}
}

@article{team2023gemini,
  title={Gemini: a family of highly capable multimodal models},
  author={Team, Gemini and Anil, Rohan and Borgeaud, Sebastian and Alayrac, Jean-Baptiste and Yu, Jiahui and Soricut, Radu and Schalkwyk, Johan and Dai, Andrew M and Hauth, Anja and Millican, Katie and others},
  journal={arXiv preprint arXiv:2312.11805},
  year={2023}
}

@article{liu2025ola,
  title={Ola: Pushing the frontiers of omni-modal language model},
  author={Liu, Zuyan and Dong, Yuhao and Wang, Jiahui and Liu, Ziwei and Hu, Winston and Lu, Jiwen and Rao, Yongming},
  journal={arXiv preprint arXiv:2502.04328},
  year={2025}
}

@article{li2025uni,
  title={{Uni-MoE-2.0-Omni}: Scaling language-centric omnimodal large model with advanced {MoE}, training and data},
  author={Li, Yunxin and Chen, Xinyu and Jiang, Shenyuan and Shi, Haoyuan and Liu, Zhenyu and Zhang, Xuanyu and Deng, Nanhao and Xu, Zhenran and Ma, Yicheng and Zhang, Meishan and others},
  journal={arXiv preprint arXiv:2511.12609},
  year={2025}
}

@article{deshmukh2026nemotron,
  title={{Nemotron 3 Nano Omni}: Efficient and open multimodal intelligence},
  author={Deshmukh, Amala Sanjay and Chumachenko, Kateryna and Rintamaki, Tuomas and Le, Matthieu and Poon, Tyler and Taheri, Danial Mohseni and Karmanov, Ilia and Liu, Guilin and Seppanen, Jarno and Goel, Arushi and others},
  journal={arXiv preprint arXiv:2604.24954},
  year={2026}
}

@article{cui2026minicpm,
  title={{MiniCPM-o 4.5}: Towards real-time full-duplex omni-modal interaction},
  author={Cui, Junbo and Xu, Bokai and Wang, Chongyi and Yu, Tianyu and Sun, Weiyue and Xu, Yingjing and Wang, Tianran and He, Zhihui and Ma, Wenshuo and Cai, Tianchi and others},
  journal={arXiv preprint arXiv:2604.27393},
  year={2026}
}

@article{gwet2008computing,

  title={Computing inter-rater reliability and its variance in the presence of high agreement},
  author={Gwet, Kilem Li},
  journal={British Journal of Mathematical and Statistical Psychology},
  volume={61},
  number={1},
  pages={29--48},
  year={2008},
  publisher={Wiley Online Library}
}

@inproceedings{avfakebench,
  title={{AVFakeBench}: A comprehensive audio-video forgery detection benchmark for {AV-LMMs}},
  author={Xia, Shuhan and Li, Peipei and Liu, Xuannan and Zhang, Dongsen and Guo, Xinyu and Li, Zekun},
  booktitle={Proceedings of the IEEE/CVF Conference on Computer Vision and Pattern Recognition},
  pages={35416--35426},
  year={2026}
}

@inproceedings{avid,
  title={{AVID}: Benchmarking Audio-Visual Inconsistency Understanding for Omni-Modal Language Models},
  author={Chen, Zixuan and Wang, Depeng and Lin, Hao and Luo, Li and Xu, Ke and Guo, Ya and Zhu, Huijia and Sun, Tanfeng and Jiang, Xinghao},
  booktitle={Proceedings of the 2026 Conference on Empirical Methods in Natural Language Processing (EMNLP)},
  year={2026}
}

@article{madbench,
  title={{MADBench}: A Benchmark for Modality-Aware Audio Deepfake Detection},
  author={Li, Yanqiu and Xiao, Yang and Bai, Jisheng and Chen, Bin and Jia, Hong and Dang, Ting},
  journal={arXiv preprint arXiv:2608.09593},
  year={2026}
}

@article{busterx,
  title={{BusterX}: {MLLM}-powered {AI}-generated video forgery detection and explanation},
  author={Wen, Haiquan and He, Yiwei and Huang, Zhenglin and Li, Tianxiao and Yu, Zihan and Huang, Xingru and Qi, Lu and Wu, Baoyuan and Li, Xiangtai and Cheng, Guangliang},
  journal={arXiv preprint arXiv:2505.12620},
  year={2025}
}

@article{deepfakebenchmm,
  title={{DeepfakeBench-MM}: A comprehensive benchmark for multimodal deepfake detection},
  author={Zhao, Kangran and Chen, Yupeng and Zhang, Xiaoyu and Chen, Yize and Guan, Weinan and Chen, Baicheng and Sun, Chengzhe and Datta, Soumyya Kanti and Liu, Qingshan and Lyu, Siwei and others},
  journal={arXiv preprint arXiv:2510.22622},
  year={2025}
}

@article{qwen25omni,
  title  = {{Qwen2.5-Omni} Technical Report},
  author = {Xu, Jin and others},
  journal = {arXiv preprint arXiv:2503.20215},
  year   = {2025}
}

@article{wan2025,
  title={Wan: Open and advanced large-scale video generative models},
  author={Wan, Team and Wang, Ang and Ai, Baole and Wen, Bin and Mao, Chaojie and Xie, Chen-Wei and Chen, Di and Yu, Feiwu and Zhao, Haiming and Yang, Jianxiao and others},
  journal={arXiv preprint arXiv:2503.20314},
  year={2025}
}

@article{seedvc,
  title={Zero-shot voice conversion with diffusion transformers},
  author={Liu, Songting},
  journal={arXiv preprint arXiv:2411.09943},
  year={2024}
}

@article{bigvgan,
  title={{BigVGAN}: A universal neural vocoder with large-scale training},
  author={Lee, Sang-gil and Ping, Wei and Ginsburg, Boris and Catanzaro, Bryan and Yoon, Sungroh},
  journal={arXiv preprint arXiv:2206.04658},
  year={2022}
}

@inproceedings{omnifake,
  title={{Omni-Fake}: Benchmarking unified multimodal social media deepfake detection},
  author={Li, Tianxiao and Huang, Zhenglin and Wen, Haiquan and He, Yiwei and Li, Xinze and Zhu, Bingyu and Duan, Wuhui and Chen, Congang and Fu, Zeyu and Dong, Yi and others},
  booktitle={Proceedings of the IEEE/CVF Conference on Computer Vision and Pattern Recognition},
  pages={30299--30311},
  year={2026}
}

@inproceedings{avlmmdetect,
  title={Leveraging large multimodal models for audio-video deepfake detection: a pilot study},
  author={Cao, Songjun and Li, Yuqi and Luo, Yunpeng and Yin, Jianjun and Ma, Long},
  booktitle={IEEE International Conference on Acoustics, Speech and Signal Processing (ICASSP)},
  pages={13442--13446},
  year={2026},
  organization={IEEE}
}

@inproceedings{shi2022avhubert,
title={Learning Audio-Visual Speech Representation by Masked Multimodal Cluster Prediction},
author={Bowen Shi and Wei-Ning Hsu and Kushal Lakhotia and Abdelrahman Mohamed},
booktitle={International Conference on Learning Representations},
year={2022},
url={https://openreview.net/forum?id=Z1Qlm11uOM}
}

@inproceedings{vyas2026pushing,
  title={Pushing the frontier of audiovisual perception with large-scale multimodal correspondence learning},
  author={Vyas, Apoorv and Chang, Heng-Jui and Yang, Cheng-Fu and Huang, Po-Yao and Gao, Luya and Richter, Julius and Chen, Sanyuan and Le, Matthew and Doll{\'a}r, Piotr and Feichtenhofer, Christoph and others},
  booktitle={Proceedings of the IEEE/CVF Conference on Computer Vision and Pattern Recognition},
  pages={30172--30182},
  year={2026}
}

@misc{facefusion,
  author       = {{FaceFusion}},
  title        = {{FaceFusion}: Next Generation Face Swapping and Enhancement},
  howpublished = {\url{https://github.com/facefusion/facefusion}},
  year         = {2026},
  note         = {GitHub repository, accessed October 28, 2026}
}

@article{xu2025qwen3omni,
  title={{Qwen3-Omni} Technical Report},
  author={Xu, Jin and others},
  journal={arXiv preprint arXiv:2509.17765},
  year={2025}
}

@article{gemma4,
  title={{Gemma 4} Technical Report},
  author={{Gemma Team}},
  journal={arXiv preprint arXiv:2607.02770},
  year={2026}
}

@article{cheng2024videollama2,
  title={{VideoLLaMA 2}: Advancing Spatial-Temporal Modeling and Audio Understanding in Video-{LLMs}},
  author={Cheng, Zesen and Leng, Sicong and Zhang, Hang and Xin, Yifei and Li, Xin and Chen, Guanzheng and Zhu, Yongxin and Zhang, Wenqi and Luo, Ziyang and Zhao, Deli and Bing, Lidong},
  journal={arXiv preprint arXiv:2406.07476},
  year={2024}
}

@inproceedings{ye2025omnivinci,
  title={{OmniVinci}: Enhancing Architecture and Data for Omni-Modal Understanding {LLM}},
  author={Ye, Hanrong and Yang, Chao-Han Huck and others},
  booktitle={International Conference on Learning Representations},
  year={2026},
  note={arXiv:2510.15870}
}

@inproceedings{fu2025vita,
  title={{VITA-1.5}: Towards {GPT-4o} Level Real-Time Vision and Speech Interaction},
  author={Fu, Chaoyou and Lin, Haojia and Wang, Xiong and Shen, Yunhang and Liu, Xiaoyu and Cao, Haoyu and Long, Zuwei and Gao, Heting and Li, Ke and MA, Long and others},
  journal={Advances in Neural Information Processing Systems},
  volume={38},
  pages={75300--75320},
  year={2026}
}

@article{peng2026havic,
  title={Leave No Stone Unturned: Uncovering Holistic Audio-Visual Intrinsic Coherence for Deepfake Detection},
  author={Peng, Jielun and Wang, Yabin and Li, Yaqi and Kong, Long and Hong, Xiaopeng},
  journal={arXiv preprint arXiv:2603.23960},
  year={2026}
}

@inproceedings{sun2026dfdhr,
  title={{DFD-HR}: Generalizable Deepfake Detection via Hierarchical Routing Learning},
  author={Sun, Jiamu and Yan, Zhiyuan and Zhang, Ke-Yue and Yao, Taiping and Ding, Shouhong},
  booktitle={Proceedings of the IEEE/CVF Conference on Computer Vision and Pattern Recognition},
  year={2026}
}

@inproceedings{deng2019arcface,
  title={{ArcFace}: Additive Angular Margin Loss for Deep Face Recognition},
  author={Deng, Jiankang and Guo, Jia and Xue, Niannan and Zafeiriou, Stefanos},
  booktitle={Proceedings of the IEEE/CVF conference on computer vision and pattern recognition},
  pages={4690--4699},
  year={2019}
}

@inproceedings{chung2016syncnet,
  title={Out of time: automated lip sync in the wild},
  author={Chung, Joon Son and Zisserman, Andrew},
  booktitle={Asian conference on computer vision},
  pages={251--263},
  year={2016},
  organization={Springer}
}

@inproceedings{desplanques2020ecapa,
  title     = {{ECAPA-TDNN: Emphasized Channel Attention, Propagation and Aggregation in TDNN Based Speaker Verification}},
  author    = {Brecht Desplanques and Jenthe Thienpondt and Kris Demuynck},
  year      = {2020},
  booktitle = {{Interspeech 2020}},
  pages     = {3830--3834},
  doi       = {10.21437/Interspeech.2020-2650},
  issn      = {2958-1796},
}

@article{raza2026deepfakes,
  title = {Deepfakes in the Foundation-Model Era: A Survey of Forensics, Generation and Distribution Across Social Media Lifecycle},
  author = {Raza, Shaina and Ho, Jessee and Radwan, Ahmed Y. and Hafez, Mohamed},
  journal = {Preprints},
  year = {2026},
  month = {August},
  publisher = {Preprints},
  doi = {10.20944/preprints202608.1870.v1},
  url = {https://doi.org/10.20944/preprints202608.1870.v1}
}

@inproceedings{cozzolino2021idreveal,
  title={{ID-Reveal}: Identity-aware DeepFake Video Detection},
  author={Cozzolino, Davide and R{\"o}ssler, Andreas and Thies, Justus and Nie{\ss}ner, Matthias and Verdoliva, Luisa},
  booktitle={2021 IEEE/CVF International Conference on Computer Vision (ICCV)},
  pages={15088--15097},
  year={2021},
  organization={IEEE}
}

@inproceedings{cozzolino2023poi,
  title={Audio-Visual Person-of-Interest DeepFake Detection},
  author={Cozzolino, Davide and Pianese, Alessandro and Nie{\ss}ner, Matthias and Verdoliva, Luisa},
  booktitle={2023 IEEE/CVF Conference on Computer Vision and Pattern Recognition Workshops (CVPRW)},
  pages={943--952},
  year={2023},
  organization={IEEE}
}

@inproceedings{lin2004rouge,
  title={{ROUGE}: A package for automatic evaluation of summaries},
  author={Lin, Chin-Yew},
  booktitle={Text Summarization Branches Out},
  pages={74--81},
  year={2004}
}

@inproceedings{papineni2002bleu,
  title={{BLEU}: a Method for Automatic Evaluation of Machine Translation},
  author={Papineni, Kishore and Roukos, Salim and Ward, Todd and Zhu, Wei-Jing},
  booktitle={Proceedings of the 40th annual meeting of the Association for Computational Linguistics},
  pages={311--318},
  year={2002}
}

@inproceedings{vedantam2015cider,
  title={{CIDEr}: Consensus-based Image Description Evaluation},
  author={Vedantam, Ramakrishna and Lawrence Zitnick, C and Parikh, Devi},
  booktitle={Proceedings of the IEEE conference on computer vision and pattern recognition},
  pages={4566--4575},
  year={2015}
}
}

\appendix
\maketitlesupplementary

\setcounter{table}{0}
\setcounter{figure}{0}
\setcounter{equation}{0}

\renewcommand{\thetable}{S\arabic{table}}
\renewcommand{\thefigure}{S\arabic{figure}}
\renewcommand{\theequation}{S\arabic{equation}}

\renewcommand{\theHtable}{supp.\arabic{table}}
\renewcommand{\theHfigure}{supp.\arabic{figure}}
\renewcommand{\theHequation}{supp.\arabic{equation}}


\begin{table}[b]
  \centering
  \scriptsize
\caption{Audio-video conditions for each source pair and retained video generator.}
  \label{tab:pair}
  \resizebox{\linewidth}{!}{%
  \begin{tabular}{lccc}
    \toprule
    & Original voice (A) & Vocoded voice (A) & Converted voice (B) \\
    \midrule
    Real video (A) & RVRA & RVRA-vocoded & RVFA \\
    Fake video (B) & FVRA & FVRA-vocoded & FVFA \\
    \bottomrule
  \end{tabular}}
\end{table}

\section{Generation and quality-check details}
\label{app:generation}

\paragraph{Clip format.} At 25\,fps and 16\,kHz, one video frame spans exactly 640 audio samples, so audio and video durations match. Both clip lengths (77 and 153 frames) contain all four authenticity classes.

\paragraph{Encoding and shortcut audit.} Every clip is re-encoded once with H.264 at CRF 18, with a common pixel format, pixel aspect ratio and container, and uncompressed audio with consistent framing. Brightness, color and loudness are normalized to common targets, and each audio track begins with 50\,ms of silence followed by a 10\,ms fade-in. We audit the result with linear probes on file, picture and sound statistics, tested on unseen identities. After normalization, the start-of-sound cue falls from AUC 0.74 to 0.57, brightness and color cues fall to 0.50, and the vocoder cue falls from 0.83 in our pilot to 0.50 on the vocoded view.
\begin{table*}[t]
  \centering
  \scriptsize
\caption{\mozaic{} composition by clip length and video generator.
RVRA and FVRA counts include original and vocoded variants.
Real-video conditions are shared across generators;
the total pair count represents unique source pairs. ``--'' denotes not applicable; real-video conditions are counted separately from generator-specific fake-video conditions.}
  \label{tab:stats}
    \resizebox{\linewidth}{!}{%
  \begin{tabular}{llrrrrrr}
    \toprule
    Length & Video method & Pairs & RVRA & RVFA & FVRA & FVFA & Clips \\
    \midrule
    \multirow{4}{*}{77f (3.08\,s)}
      & Real video       & 7,590 & 15,180 & 7,590 & -- & -- & 22,770 \\
      & FaceFusion       & 7,020 & -- & -- & 14,040 & 7,020 & 21,060 \\
      & Wan2.2-S2V       & 6,360 & -- & -- & 12,720 & 6,360 & 19,080 \\
      & Wan2.2-Animate   & 4,407 & -- & -- & 8,814 & 4,407 & 13,221 \\
    \midrule
    \multirow{3}{*}{153f (6.12\,s)}
      & Real video       & 7,424 & 14,848 & 7,424 & -- & -- & 22,272 \\
      & FaceFusion       & 6,835 & -- & -- & 13,670 & 6,835 & 20,505 \\
      & Wan2.2-S2V       & 6,279 & -- & -- & 12,558 & 6,279 & 18,837 \\
    \midrule
    \multicolumn{2}{l}{\textbf{Total}}
      & \textbf{15,014} & \textbf{30,028} & \textbf{15,014}
      & \textbf{61,802} & \textbf{30,901} & \textbf{137,745} \\
      \midrule
\multicolumn{8}{l}{\emph{Splits (IDs / pairs / clips):}
train 700 / 10,541 / 96,957 \quad
val 150 / 2,232 / 20,553 \quad
test 150 / 2,241 / 20,235} \\

\multicolumn{8}{l}{\mozaicsmall{} (pairs / clips):
train 1,087 / 9,996 \quad
val 217 / 1,998 \quad
test 554 / 5,001 \quad total 1,858 / 16,995} \\
    \bottomrule
    
  \end{tabular}
  }
\end{table*}

\begin{figure}[h]
  \centering
  \includegraphics[width=0.75\linewidth]{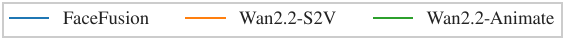}\\[2pt]
  \begin{minipage}{0.49\linewidth}\centering
    \includegraphics[width=\linewidth]{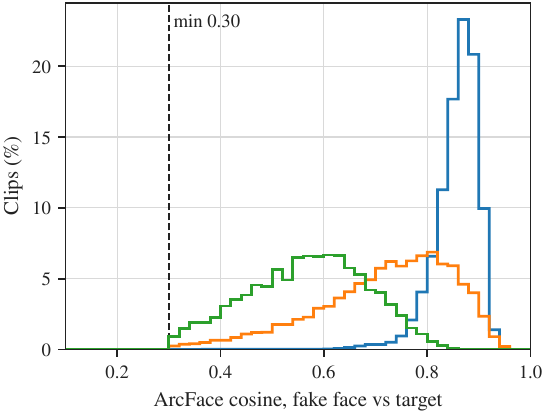}\\{\footnotesize (a)}
  \end{minipage}\hfill
  \begin{minipage}{0.49\linewidth}\centering
    \includegraphics[width=\linewidth]{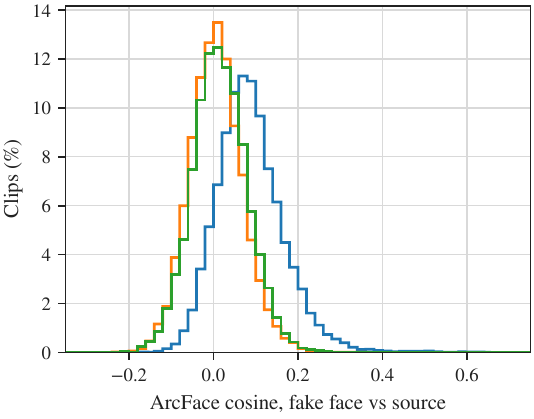}\\{\footnotesize (b)}
  \end{minipage}\\[4pt]
  \begin{minipage}{0.49\linewidth}\centering
    \includegraphics[width=\linewidth]{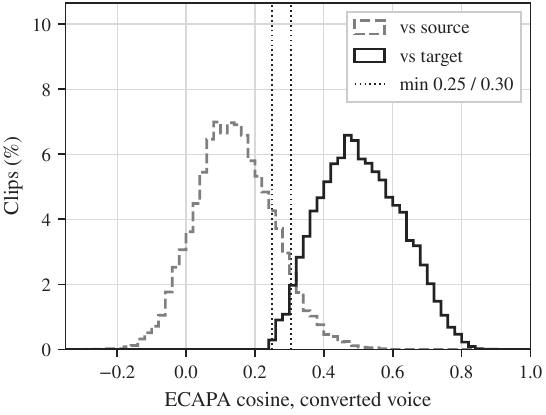}\\{\footnotesize (c)}
  \end{minipage}\hfill
  \begin{minipage}{0.49\linewidth}\centering
    \includegraphics[width=\linewidth]{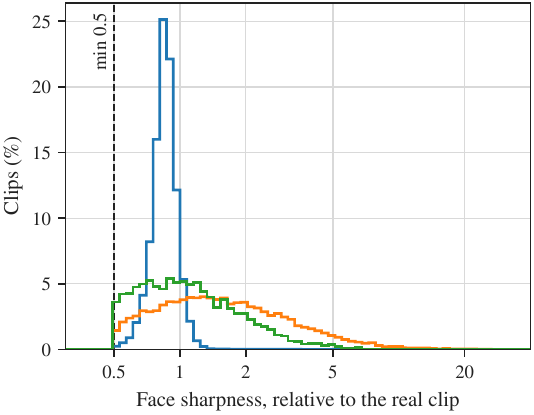}\\{\footnotesize (d)}
  \end{minipage}
  \caption{Quality-check scores of the clips per generator. (a,b)~ArcFace similarity of the fake face to the target B and the source A; (c)~converted voice vs target and source (ECAPA-TDNN~\cite{desplanques2020ecapa}); (d)~face sharpness relative to the real clip. Dashed lines: thresholds; in (c), dotted lines: the voice limit for 3\,s and 6\,s clips.}
  \label{fig:stats}
\end{figure}

\begin{figure*}[t]
  \centering
  \includegraphics[width=\textwidth]{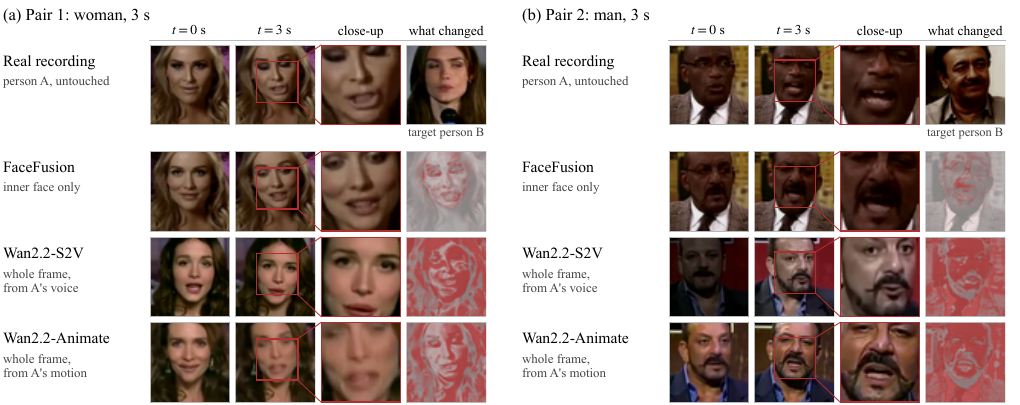}
  \caption{\textbf{What each generator changes, two test pairs.} Within a pair, the four rows start from the same real frames of person A and differ only by the generator. FaceFusion replaces the inner face and keeps A's hair, body and background; Wan2.2-Animate and Wan2.2-S2V redraw the whole frame, so the scene becomes B's rather than A's. The red box and call-out give a $2\times$ close-up of the face at $t=3$\,s. Last column: the real frame in grey with every pixel the generator changed in red; in the real row it shows the target person B instead. 
  }
  \label{fig:qualitative_full}
\end{figure*}

FaceFusion uses \texttt{inswapper\_128\_fp16}. Wan2.2-S2V-14B runs at 480p with 40 steps and CFG 4.5, renders at 16\,fps and is retimed to 25\,fps. Wan2.2-Animate-14B runs in replace mode at 480p. Seed-VC uses 5 steps and CFG 0.3, with one reference recording of the target, and BigVGAN v2 (22\,kHz) renders the converted and vocoded voices. Pairs were drawn with seed 42 and rendered on A100 and A40 GPUs in 40 batches of 200 pairs per length. Two checks are switched off for one generator each: the limit on remaining source-face similarity for FaceFusion, which keeps the source's hair and head by design, and the melting check for Wan2.2-S2V, which invents its own head movement. \Cref{tab:rejections} gives the removal reasons.

\begin{table*}[t]
  \centering
  \scriptsize
  \caption{Pairs failing each check (\% of 8,000 candidates). A pair can fail several checks. Incomplete: a required clip is missing, including after a refusal. Any: failed at least one check in the final selection reports. Released: removed from the release; it differs from Any by at most 0.3 points.}
  \label{tab:rejections}
  \resizebox{\linewidth}{!}{%
  \begin{tabular}{lccccccccc}
    \toprule
    Generator & Length & Incomplete & Refused & Face & Voice & Lip sync & Picture & Any & Released \\
    \midrule
    FaceFusion & 3\,s & 6.2 & 5.1 & 0.8 & 3.3 & 1.6 & 0.8 & 12.5 & 12.3 \\
    FaceFusion & 6\,s & 7.7 & 6.7 & 0.7 & 4.0 & 1.9 & 0.9 & 14.9 & 14.6 \\
    Wan2.2-S2V & 3\,s & 1.0 & 0.0 & 2.8 & 3.5 & 1.4 & 12.9 & 20.3 & 20.5 \\
    Wan2.2-S2V & 6\,s & 1.4 & 0.0 & 3.7 & 4.4 & 1.7 & 11.6 & 21.5 & 21.5 \\
    Wan2.2-Animate & 3\,s & 2.1 & 0.0 & 12.0 & 3.5 & 3.6 & 30.9 & 44.9 & 44.9 \\
    \bottomrule
  \end{tabular}}
\end{table*}

\begin{figure*}[t]
    \centering
    \includegraphics[width=\textwidth]{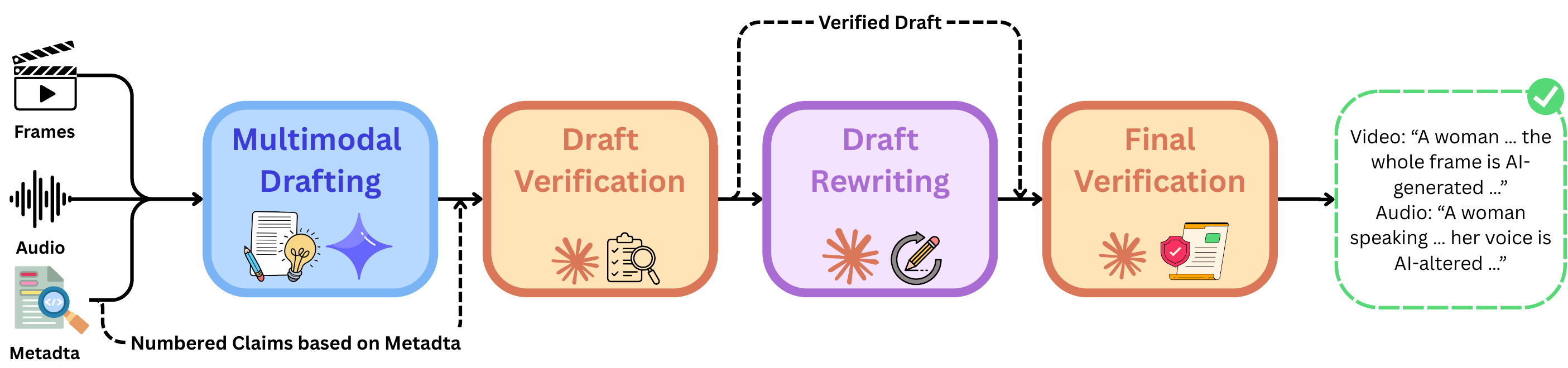}
    \caption{\textbf{AI-assisted construction of T4 reference explanations.} Gemini drafts explanations from video frames, audio, and metadata-derived claims. Claude verifies the draft against these claims, rewrites the checked sentences,and verifies the rewritten text against the verified draft. Both verification stages include automated checks and text-only model review. Dashed arrows indicate reference inputs.}
    \label{fig:t4-annotation-pipeline}
\end{figure*}

\section{AI-Assisted Annotation Process}
\label{app:t4-annotation}

\subsection{Model-assisted Drafting}

We expand \mozaicsmall{} with T4, open-ended explanations that describe the video and audio separately, identifying whether each modality is real or AI-generated and, when manipulated, what was changed. Reference construction follows four stages: multimodal drafting, draft verification, rewriting, and final verification. Ground-truth authenticity labels and manipulation regions are derived from the dataset metadata, while the models express these facts in natural language and describe the clip's content.

\paragraph{Multimodal drafting.}
We use Gemini 3.6 Flash to produce initial drafts from eight uniformly sampled video frames and the clip's audio. The model also receives the person's gender from metadata and a set of numbered claims specifying each modality's authenticity, the manipulated region, and any supported visual signs identified through quality assessment. For example, FaceFusion clips are associated with inner-face manipulation, whereas Wan2.2-Animate and Wan2.2-S2V clips are associated with whole-frame generation; converted audio is associated with voice alteration. Gemini describes the scene and speech content while grounding its manipulation descriptions in the supplied claims, with each draft sentence citing the claims it uses. Real and manipulated modalities follow the same annotation stages with separate instructions: real modalities are described in terms of their content and explicitly identified as natural recordings. The annotator prompts are provided in \cref{annotation-gemini}.

\paragraph{Draft verification.}
Automated checks first validate the response structure, sentence counts, claim citations, and coverage of the required facts. Claude Sonnet 5.5 then reviews the draft sentences against their cited claims using text only, checking for unsupported manipulation descriptions, quality claims, and added judgments (\cref{annotation-verifier}). This review assesses consistency with the supplied claims; it does not independently inspect the clip. Drafts that fail either check are returned to Gemini with feedback for revision.

\paragraph{Rewriting.}
Once a draft passes verification, Claude Sonnet 5.5 rewrites the checked sentences into one coherent paragraph per modality (\cref{annotation-rewriter}). The rewriter receives only the checked text and is instructed to preserve the scene description, authenticity verdict, manipulation region, and stated signs while improving flow and removing repetition. It must not introduce new scene details or manipulation claims.

\paragraph{Final verification.}
The rewritten paragraphs undergo a second set of automated checks and a further text-based review by Claude Sonnet 5.5 (\cref{annotation-rewrite-verifier}). This pass compares the rewrite with the previously checked sentences, checking for newly introduced details and changes to the meaning or strength of the stated evidence. Rewrites that fail are revised and checked again. Only explanations that satisfy both verification stages are retained.

\paragraph{Revision statistics.}
The pipeline records real and manipulated parts separately for each clip; a clip containing one real and one manipulated modality therefore contributes two annotation records. Across the 16,995 retained records, 12,354 (72.7\%) passed draft verification in the first round and 4,641 (27.3\%) passed in the second round after feedback. At the rewriting stage, 344 records (2.0\%) were successfully corrected through retries. All 16,995 records passed the final quality checks, with no unresolved failures in the saved results. These statistics summarize the retained annotation records rather than all attempts made during annotation.

\paragraph{Human validation.}
Three of the authors independently assessed a stratified set of 400 clips (150 FVFA, 150 FVRA, 50 RVFA, and 50 RVRA), rating the video and audio references separately. Majority judgments classify 99.2\% of references as acceptable, with high inter-rater agreement (Gwet's AC1~\cite{gwet2008computing}: 0.985 for video and 0.983 for audio); we use AC1 because positive judgments dominate. The reviewers rated the references before the final rewording pass, which kept their facts. The raters worked on model-written text only, so no participants were recruited and no ethics review was required.

\subsection{AI-Generation Prompts}
\label{app:t4-generation-prompts}

The following templates are used to draft, verify, and rewrite the T4 reference explanations. Placeholders are populated with clip-specific claims or intermediate text. Separate annotator and draft-verifier prompts are used for real and manipulated modalities.

\subsubsection{Annotator}
\label{annotation-gemini}

Gemini 3.6 Flash receives the clip's eight sampled frames and audio before the prompt. Here, \texttt{\{person\}} supplies the metadata-derived person description, \texttt{\{blocks\}} contains the numbered claims and requested sentence counts for the relevant modalities, and \texttt{\{reply\}} specifies a JSON response containing an array of sentences per modality, each with \texttt{text} and \texttt{claims} fields.

\paragraph{Manipulated parts.}

\begin{tcolorbox}[
    breakable, width=\linewidth,
    colback=blue!5!white, colframe=blue!75!black,
    title={T4 Annotation: Annotator Prompt (Manipulated Parts)},
    fonttitle=\bfseries, fontupper=\scriptsize,
    boxrule=0.5mm, arc=2mm
]
\begin{lstlisting}[
    breaklines=true,
    basicstyle=\ttfamily\scriptsize,
    columns=fullflexible
]
You are writing the reference explanation for one clip of a deepfake benchmark. The clip is shown above. For each part, write one flowing paragraph, the way a good narrator would explain it to someone watching with you: set the scene (who is there, where, what they do or say), then explain what is AI-generated and for how long, and how the listed signs show as the clip plays. Use full, connected sentences that build on each other, not a list of short statements, and follow the clip as it unfolds (what happens first, then next).

The numbered claims below are the facts about this clip. Cite the claims each sentence uses, and use every one ("what is on screen" and "what is heard" are there to cite whenever you describe the scene). You may describe what you see and hear freely and neutrally, and say where or when a listed sign shows. But every defect, flaw or oddity you mention must be one the claims list: never add another, never say anything looks natural, real or convincing, never name how it was made (a method, a tool, "deepfake"), never compare it with real footage, and don't talk about judging or concluding.

The person in the clip is a {person}. Cite claims only in "claims", never in the text, and never give a time, a timestamp or a frame number: the times printed on the frames are only there to help you follow the clip.

{blocks}

Reply in JSON only: {reply}
\end{lstlisting}
\end{tcolorbox}

\paragraph{Real parts.}

\begin{tcolorbox}[
    breakable, width=\linewidth,
    colback=blue!5!white, colframe=blue!75!black,
    title={T4 Annotation: Annotator Prompt (Real Parts)},
    fonttitle=\bfseries, fontupper=\scriptsize,
    boxrule=0.5mm, arc=2mm
]
\begin{lstlisting}[
    breaklines=true,
    basicstyle=\ttfamily\scriptsize,
    columns=fullflexible
]
You are writing the reference explanation for one clip of a deepfake benchmark. The clip is shown above. Every part listed below is a natural recording: nothing in it is AI-generated, and there is nothing to find or prove. For each part, write one flowing paragraph that describes the clip, the way a good narrator would  describe it to someone watching with you: for VIDEO, who is there, where, and what they do; for AUDIO, who speaks and what they say. Follow the clip as it unfolds (what happens first, then next), in full, connected sentences that build on each other, not a list of short statements. End the paragraph by saying that the part is a natural recording, with nothing AI-generated, in those words.

Describe what is there, never its quality: not the face's expressions, detail, sharpness or how it moves, not how the light falls, not how the voice sounds (its tone, pace, clarity or smoothness). Never mention a defect, flaw or oddity, never call anything natural, real or convincing except in the last sentence, never name how a clip is made or processed (a method, a tool, a vocoder, compression), never compare it with other footage, and don't talk about judging or concluding (no "overall", no "concludes").

The numbered claims below are the facts about this clip. Cite the claims each sentence uses, and use every one ("what is on screen" and "what is heard" are there to cite whenever you describe the scene). Each paragraph is about its own part only: the picture for VIDEO, the sound for AUDIO.

The person in the clip is a {person}. Cite claims only in "claims", never in the text, and never give a time, a timestamp or a frame number: the times printed on the frames are only there to help you follow the clip.

{blocks}

Reply in JSON only: {reply}
\end{lstlisting}
\end{tcolorbox}

\paragraph{Revision feedback.}
When a draft fails the checks, the following feedback is appended to the corresponding annotator prompt:

\begin{tcolorbox}[
    breakable, width=\linewidth,
    colback=blue!5!white, colframe=blue!75!black,
    title={T4 Annotation: Revision Feedback},
    fonttitle=\bfseries, fontupper=\scriptsize,
    boxrule=0.5mm, arc=2mm
]
\begin{lstlisting}[
    breaklines=true,
    basicstyle=\ttfamily\scriptsize,
    columns=fullflexible
]
Your last reply failed these checks; write it again:
- {problems}
\end{lstlisting}
\end{tcolorbox}

\subsubsection{Draft Verifier}
\label{annotation-verifier}

Claude Sonnet 5.5 receives the numbered claims and the draft sentences with their claim citations. It does not receive the clip. The response contains one verification decision per sentence.

\paragraph{Manipulated parts.}

\begin{tcolorbox}[
    breakable, width=\linewidth,
    colback=orange!5!white, colframe=orange!50!black,
    title={T4 Annotation: Draft Verifier (Manipulated Parts)},
    fonttitle=\bfseries, fontupper=\scriptsize,
    boxrule=0.5mm, arc=2mm
]
\begin{lstlisting}[
    breaklines=true,
    basicstyle=\ttfamily\scriptsize,
    columns=fullflexible
]
Below are claims about a video clip (C1, C2, ...), and sentences written from them (S1, S2, ...) by someone who also watched the clip. Each sentence lists the claims it is based on.

The writer may describe what is on screen neutrally (who, where, what they do), and may say where or when a listed defect shows. 
A sentence fails if it states or implies any defect, flaw or oddity that no listed claim gives, if it says anything looks or sounds natural, real, fine or convincing, or if it adds a cause or a judgement. It also fails if it has claim numbers in its text (such as [3] or (2)), a timestamp, a time in seconds, or a frame number. Rewording, combining and linking the listed claims is fine.

Claims:
{claims}

Sentences:
{sentences}

Reply in JSON only, one entry per sentence: {"S1": {"follows": true or false, "added": "<what it adds, or empty>"}, ...}
\end{lstlisting}
\end{tcolorbox}

\paragraph{Real parts.}

\begin{tcolorbox}[
    breakable, width=\linewidth,
    colback=orange!5!white, colframe=orange!50!black,
    title={T4 Annotation: Draft Verifier (Real Parts)},
    fonttitle=\bfseries, fontupper=\scriptsize,
    boxrule=0.5mm, arc=2mm
]
\begin{lstlisting}[
    breaklines=true,
    basicstyle=\ttfamily\scriptsize,
    columns=fullflexible
]
Below are claims about a video clip (C1, C2, ...), and sentences written from them (S1, S2, ...) by someone who also watched the clip. Each sentence lists the claims it is based on. Every part the sentences are about is a natural recording.

The writer may describe what is on screen or heard neutrally (who, where, what they do or say), and says the part is a natural recording, with nothing AI-generated. 
A sentence fails if it states or implies any defect, flaw, oddity or sign of AI generation or editing, if it describes the quality of the picture or the sound (how a face, its expressions or the light look, how the voice sounds: its tone, pace, clarity or smoothness), if it says how natural, real or convincing anything looks or sounds beyond that claim, or if it adds a cause or a judgement. It also fails if it has claim numbers in its text (such as [3] or (2)), a timestamp, a time in seconds, or a frame number. Rewording, combining and linking the listed claims is fine.

Claims:
{claims}

Sentences:
{sentences}

Reply in JSON only, one entry per sentence: {"S1": {"follows": true or false, "added": "<what it adds, or empty>"}, ...}
\end{lstlisting}
\end{tcolorbox}
\subsubsection{Rewriter}
\label{annotation-rewriter}

Claude Sonnet 5.5 receives the checked sentences as text and rewrites each modality as one paragraph. Both real and manipulated parts use the same system prompt. The user prompt for real parts includes an additional instruction to prevent unsupported details about posture or actions. Only the modalities included in the current call appear in the message and response.

\paragraph{System prompt.}

\begin{tcolorbox}[
    breakable, width=\linewidth,
    colback=green!5!white, colframe=green!50!black,
    title={T4 Annotation: Rewriter System Prompt},
    fonttitle=\bfseries, fontupper=\scriptsize,
    boxrule=0.5mm, arc=2mm
]
\begin{lstlisting}[
    breaklines=true,
    basicstyle=\ttfamily\scriptsize,
    columns=fullflexible
]
Rewrite the sentences of each part of a clip (the video, the audio) as one paragraph, the way a person would write it for someone watching the clip with them.

Say what the sentences say and nothing more: the same people, actions and facts, the words of each flaw exactly as they are, and "AI-generated" and "AI-altered" wherever they are used. Only the wording and the order change.

Write plainly and naturally, stating facts without contrast or drama: say each point once, let sentences of different lengths flow into each other, and use no filler, stock phrases, dashes, lists, bracketed numbers or times (no timestamps or seconds).
\end{lstlisting}
\end{tcolorbox}

\paragraph{User prompt for manipulated parts.}

\begin{tcolorbox}[
    breakable, width=\linewidth,
    colback=green!5!white, colframe=green!50!black,
    title={T4 Annotation: Rewriter User Prompt (Manipulated Parts)},
    fonttitle=\bfseries, fontupper=\scriptsize,
    boxrule=0.5mm, arc=2mm
]
\begin{lstlisting}[
    breaklines=true,
    basicstyle=\ttfamily\scriptsize,
    columns=fullflexible
]
VIDEO:
<checked sentences>

AUDIO:
<checked sentences>

Reply in JSON only: {"video": "<paragraph>", "audio": "<paragraph>"}
\end{lstlisting}
\end{tcolorbox}

\paragraph{User prompt for real parts.}

\begin{tcolorbox}[
    breakable, width=\linewidth,
    colback=green!5!white, colframe=green!50!black,
    title={T4 Annotation: Rewriter User Prompt (Real Parts)},
    fonttitle=\bfseries, fontupper=\scriptsize,
    boxrule=0.5mm, arc=2mm
]
\begin{lstlisting}[
    breaklines=true,
    basicstyle=\ttfamily\scriptsize,
    columns=fullflexible
]
VIDEO:
<checked scene sentences>

AUDIO:
<checked scene sentences>

Keep every detail exactly as the sentences give it: where they say a person is positioned somewhere, do not say they stand or sit, and add no posture, direction or action they do not state.

Reply in JSON only: {"video": "<paragraph>", "audio": "<paragraph>"}
\end{lstlisting}
\end{tcolorbox}

\subsubsection{Final Verifier}
\label{annotation-rewrite-verifier}

Claude Sonnet 5.5 receives the checked sentences as its reference and the rewritten sentences as the text to assess. The same template is used for real and manipulated modalities. This pass checks whether rewriting introduces unsupported details or changes the meaning or strength of the stated evidence.

\begin{tcolorbox}[
    breakable, width=\linewidth,
    colback=purple!5!white, colframe=purple!50!black,
    title={T4 Annotation: Final Verification Prompt},
    fonttitle=\bfseries, fontupper=\scriptsize,
    boxrule=0.5mm, arc=2mm
]
\begin{lstlisting}[
    breaklines=true,
    basicstyle=\ttfamily\scriptsize,
    columns=fullflexible
]
Below are checked sentences about a video clip (C1, C2, ...), and a rewrite of them (S1, S2, ...) by someone who did not see the clip. Each sentence of the rewrite lists the sentences it is based on.

A sentence of the rewrite fails if it states anything the listed sentences do not: a detail of the scene they do not give (posture, place, clothing, an action, what is said), a defect, flaw or oddity, a cause, a judgement, or a flaw said more or less strongly. It also fails if it has claim numbers in its text (such as [3] or (2)), a timestamp, a time in seconds, or a frame number. Rewording, combining, reordering, linking points (with and, so, as) and leaving out repeats is fine.

Sentences:
{claims}

Rewrite:
{sentences}

Reply in JSON only, one entry per sentence: {"S1": {"follows": true or false, "added": "<what it adds, or empty>"}, ...}
\end{lstlisting}
\end{tcolorbox}

\section{Task-Specific Prompts}
\label{app:prompts}

\begin{table}[t]
  \centering
  \scriptsize
  \caption{Evaluation tasks and question counts across training, validation, and test splits. T2 uses video only, while T4 is evaluated on \mozaicsmall{}.}
  \label{tab:tasks}
  \resizebox{\linewidth}{!}{%
  \begin{tabular}{llrr}
    \toprule
    Task & Input / Output & \mozaic{} & \mozaicsmall{} \\
    \midrule
    T1 Authenticity & V+A / 4-way & 64,638 / 13,702 / 13,490 & 6,664 / 1,332 / 3,334 \\
    T2 Manipulation & V / 4-way & 32,319 / 6,851 / 6,745 & 3,332 / 666 / 1,667 \\
    T3 Identity & 2 clips / 4-way & 106,802 / 22,586 / 22,454 & 11,012 / 2,200 / 5,550 \\
    T4 Explanation & V+A / free text & -- & 6,664 / 1,332 / 3,334 \\
    \bottomrule
  \end{tabular}}
\end{table}

One template per task is used for every model, at temperature 0. The question, options and descriptions below are filled in as shown, with \texttt{\{n\}} the number of frames; options and descriptions follow each question's shuffled order. In the modality ablation only the first line of the T1 prompt changes.

\subsection{T1: Audio-video authenticity prompt}

\begin{tcolorbox}[
    breakable,
    width=\linewidth,
    colback=blue!5!white,
    colframe=blue!75!black,
    title=T1: Audio-video Authenticity,
    fonttitle=\bfseries,
    fontupper=\scriptsize,
    boxrule=0.5mm,
    arc=2mm
]
\begin{lstlisting}[
    breaklines=true,
    basicstyle=\ttfamily\scriptsize,
    columns=fullflexible
]
The images are {n} frames sampled evenly from one video clip, in time order. The audio is that clip's soundtrack.

Which part of this audio-visual clip, if any, is AI-generated or manipulated? A real recording that was only processed, without changing who is seen or heard, still counts as real.

What each option means:
- "the video": the video is AI-generated or edited; the audio is real
- "the audio": the audio is AI-generated or edited; the video is real
- "both": both the video and the audio are AI-generated or edited
- "neither": both the video and the audio are real

Options: "the video", "the audio", "both", "neither"

Reply in JSON only: {"answer": "<one of the options, copied exactly>"}
\end{lstlisting}
\end{tcolorbox}

\noindent\textbf{First line without sound:}
\emph{The images are \{n\} frames sampled evenly from one video clip, in time order. The clip's soundtrack is not provided.}

\medskip
\noindent\textbf{First line without frames:}
\emph{The audio is one video clip's soundtrack. The clip's frames are not provided.}

\subsection{T2: Manipulation type prompt}

\begin{tcolorbox}[
    breakable,
    width=\linewidth,
    colback=orange!5!white,
    colframe=orange!50!black,
    title=T2: Manipulation Type,
    fonttitle=\bfseries,
    fontupper=\scriptsize,
    boxrule=0.5mm,
    arc=2mm
]
\begin{lstlisting}[
    breaklines=true,
    basicstyle=\ttfamily\scriptsize,
    columns=fullflexible
]
The images are {n} frames sampled evenly from one video clip, in time order. There is no sound.

Which part of this video, if any, was generated or altered by AI? Ignore normal processing such as brightness, colour or compression.

What each option means:
- "real": a camera recording; nothing was generated
- "face swap": only the inner face was replaced; hair, head shape, body, background and all movement are from the original recording
- "fully generated": the whole frame was generated, including the person and the background
- "full-body swap": the whole frame was redrawn from one photo of another person, copying the movement of the original recording

Options: "real", "face swap", "fully generated", "full-body swap"

Reply in JSON only: {"answer": "<one of the options, copied exactly>"}
\end{lstlisting}
\end{tcolorbox}

\subsection{T3: Identity verification, face questions prompt}
The two clips are sent first, labeled \emph{Clip 1:} and \emph{Clip 2:}.

\begin{tcolorbox}[
    breakable,
    width=\linewidth,
    colback=purple!5!white,
    colframe=purple!50!black,
    title={T3: Identity Verification, Face Questions},
    fonttitle=\bfseries,
    fontupper=\scriptsize,
    boxrule=0.5mm,
    arc=2mm
]
\begin{lstlisting}[
    breaklines=true,
    basicstyle=\ttfamily\scriptsize,
    columns=fullflexible
]
Above are {n} frames sampled evenly from each clip, in time order: first clip 1, then clip 2. There is no sound.

Clip 2 is a real recording of one person. Look only at the faces (the inner face: eyes, nose and mouth). Is the face in clip 1 real or fake, and is it the same person as in clip 2? A real recording that was only processed, without changing whose face it is, still counts as real.

What each option means:
- "real, same person": the face in clip 1 is a real recording of the person in clip 2
- "real, different person": the face in clip 1 is a real recording of someone else
- "fake of this person": the face in clip 1 was swapped, changed or made by a computer to look like the person in clip 2
- "fake of a different person": the face in clip 1 was swapped, changed or made by a computer, and looks like someone else

Options: "real, same person", "real, different person", "fake of this person", "fake of a different person"

Reply in JSON only: {"answer": "<one of the options, copied exactly>"}
\end{lstlisting}
\end{tcolorbox}

\subsection{T3: Identity verification, voice questions prompt}

\begin{tcolorbox}[
    breakable,
    width=\linewidth,
    colback=purple!5!white,
    colframe=purple!50!black,
    title={T3: Identity Verification, Voice Questions},
    fonttitle=\bfseries,
    fontupper=\scriptsize,
    boxrule=0.5mm,
    arc=2mm
]
\begin{lstlisting}[
    breaklines=true,
    basicstyle=\ttfamily\scriptsize,
    columns=fullflexible
]
Above are the soundtracks of clip 1 and clip 2, in that order. There are no images.

Clip 2 is a real recording of one person. Listen only to the voices. Is the voice in clip 1 real or fake, and is it the same person as in clip 2? A real recording that was only processed, without changing whose voice it is, still counts as real.

What each option means:
- "real, same person": the voice in clip 1 is a real recording of the person in clip 2
- "real, different person": the voice in clip 1 is a real recording of someone else
- "fake of this person": the voice in clip 1 was converted, changed or made by a computer to sound like the person in clip 2
- "fake of a different person": the voice in clip 1 was converted, changed or made by a computer, and sounds like someone else

Options: "real, same person", "real, different person", "fake of this person", "fake of a different person"

Reply in JSON only: {"answer": "<one of the options, copied exactly>"}
\end{lstlisting}
\end{tcolorbox}

\subsection{T4: Evidence-based explanation prompt}

\begin{tcolorbox}[
    breakable,
    width=\linewidth,
    colback=green!5!white,
    colframe=green!50!black,
    title=T4: Evidence-based Explanation,
    fonttitle=\bfseries,
    fontupper=\scriptsize,
    boxrule=0.5mm,
    arc=2mm
]
\begin{lstlisting}[
    breaklines=true,
    basicstyle=\ttfamily\scriptsize,
    columns=fullflexible
]
The images are {n} frames sampled evenly from one video clip, in time order. The audio is that clip's soundtrack.

Some clips are real recordings; normal processing such as compression does not count as AI generation. In others, AI generated or altered the video, the audio, or both.

Explain the video and the audio of this clip, each on its own, in 3 to 5 full sentences per part:
- video: say clearly whether the video is AI-generated: if so, what was generated; if not, that it is a natural recording, with nothing AI-generated.
- audio: say clearly whether the audio is AI-generated: if so, what was generated or altered; if not, that it is a natural recording, with nothing AI-generated.
Describe what is on screen (who is shown, where, what they do) and what is heard (who speaks and what about), and name any visible sign of manipulation.

What can be generated or altered:
- inner face: only the inner face (eyes, nose, mouth, skin) is AI-generated; the hair, head, body and background are untouched
- whole frame: the whole frame, person and background, is AI-generated
- voice: the voice is AI-altered

Reply in JSON only, each value a paragraph: {"video": "<3 to 5 sentences>", "audio": "<3 to 5 sentences>"}
\end{lstlisting}
\end{tcolorbox}

\subsection{T4: Judge prompt}
The judge (\texttt{gpt-5.4-mini}) sees text only, never the clip.

\begin{tcolorbox}[
    breakable,
    width=\linewidth,
    colback=green!5!white,
    colframe=green!50!black,
    title=T4 Judge: System Prompt,
    fonttitle=\bfseries,
    fontupper=\scriptsize,
    boxrule=0.5mm,
    arc=2mm
]
\begin{lstlisting}[
    breaklines=true,
    basicstyle=\ttfamily\scriptsize,
    columns=fullflexible
]
You are an intelligent and fair evaluator of explanations of audio-video clips that may be AI-generated or altered. For one clip you get a reference explanation and a predicted explanation, each with a video part and an audio part. Score each predicted part against the same reference part.

Evaluation instructions:
- Judge meaning, not wording: accept paraphrases, synonyms and differences in style or tone.
- Ignore timing words: every manipulation covers the whole clip.
- For each part, weigh in this order of importance:
  1. Fake or real: the part called AI-generated/altered, or a real recording, as in the reference. A hedged or
     self-contradictory answer ("may be AI-generated", "real, but parts look generated") counts as wrong.
  2. What was manipulated: for a fake video, the same region as the reference: only the inner face (eyes, nose, mouth, skin) with the rest untouched, or the whole frame (person and background).
     For a fake audio, that the voice was generated or altered.
  3. Reasonableness: nothing called AI-generated that the reference says is real (a real part, or the background or body when only the inner face was generated). Visible signs the reference names (e.g. the face looks blurry, warps, barely moves) are credited when matched; signs the reference does not name are neither rewarded nor penalized.
  4. Scene: a description of who is shown or heard, where, and what about that agrees with the reference is fine; one that contradicts it (a woman instead of a man, another setting or topic) costs a little; a missing description costs nothing.
- When uncertain between two scores, give the lower one.

Scale for each part (0-10):
  10   fake or real as in the reference; for a fake part, what was manipulated right and the reference's
       signs (if any) named; nothing hallucinated. A real part called real with nothing hallucinated is 10.
  8-9  fake or real, and what was manipulated, right; a minor gap (a sign missed, a small scene contradiction)
  5-7  fake or real right, but what was manipulated is wrong or unstated, or more is called manipulated
       than the reference says
  1-4  a fake part called real, or a real part called fake
  0    no usable explanation

Return a JSON object only:
{"video": {"score": <integer 0-10>, "justification": "<concise explanation (1-2 sentences)>"},
 "audio": {"score": <integer 0-10>, "justification": "<concise explanation (1-2 sentences)>"}}
\end{lstlisting}
\end{tcolorbox}

\begin{tcolorbox}[
    breakable,
    width=\linewidth,
    colback=green!5!white,
    colframe=green!50!black,
    title=T4 Judge: User Prompt,
    fonttitle=\bfseries,
    fontupper=\scriptsize,
    boxrule=0.5mm,
    arc=2mm
]
\begin{lstlisting}[
    breaklines=true,
    basicstyle=\ttfamily\scriptsize,
    columns=fullflexible
]
Please evaluate the following explanation of an audio-video clip.
Reference
  video: {gt_video}
  audio: {gt_audio}
Prediction
  video: {pred_video}
  audio: {pred_audio}
Please return your evaluation in the specified JSON format, with a score and a justification for each part.
\end{lstlisting}
\end{tcolorbox}

\section{Per-model settings}
\label{app:models}
\Cref{tab:models} lists each model's input path. Every model receives 32 frames per clip, sampled by its own preprocessing when it takes video natively and evenly spaced otherwise, and 16\,kHz sound. OmniVinci averages every 8 neighboring frames inside the model, so it uses 4 pictures per clip.

\begin{table*}[t]
  \centering
  \scriptsize
  \caption{Model settings. Sound: inside the video, or sent as a separate part. Frame size: pixels the model's picture encoder receives; \emph{own} means the model's own preprocessing sets it, and Gemini runs at its low video resolution (64 tokens per frame). GPU refers to the T1 run on the \mozaic{} test set. VideoLLaMA2.1-AV is its nopad run for T1, T3 and T4.}
  \label{tab:models}
  \resizebox{\linewidth}{!}{%
  \begin{tabular}{lllcccl}
    \toprule
    Model & Weights / API name & Engine & Sound & Frames used & Frame size & GPU \\
    \midrule
    Qwen3-Omni & \texttt{Qwen/Qwen3-Omni-30B-A3B-Instruct} & vLLM 0.17.1 & in video & 32 & 224 & A40 \\
    Qwen2.5-Omni & \texttt{Qwen/Qwen2.5-Omni-7B} & vLLM 0.17.1 & in video & 32 & 224 & A40 \\
    MiniCPM-o 4.5 & \texttt{openbmb/MiniCPM-o-4\_5} & vLLM 0.28.0 & separate & 32 & own & A100 \\
    Nemotron 3 Nano Omni & \texttt{nvidia/Nemotron-3-Nano-Omni-30B-A3B-Reasoning-BF16} & vLLM 0.28.0 & in video & 32 & 224 & A40 \\
    Gemma 4 & \texttt{google/gemma-4-12B-it} & vLLM 0.28.0 & separate & 32 & 384 & A40 \\
    VideoLLaMA2.1-AV & \texttt{DAMO-NLP-SG/VideoLLaMA2.1-7B-AV} & Transformers 4.42.3 & in video & 32 & 384 & A40 \\
    OmniVinci & \texttt{nvidia/omnivinci} & Transformers 4.46.0 & in video & 4 (from 32) & 448 & A40 \\
    Uni-MoE-2.0-Omni & \texttt{HIT-TMG/Uni-MoE-2.0-Omni} & Transformers 4.51.3 & separate & 32 & 384 & A100 \\
    Ola & \texttt{THUdyh/Ola-7b} & Transformers 4.43.4 & separate & 32 & 256 & A100 \\
    VITA-1.5 & \texttt{VITA-MLLM/VITA-1.5} & Transformers 4.44.2 & separate & 32 & 448 & A40 \\
    \midrule
    Gemini 3.8 Flash & \texttt{gemini-3.8-flash} & API & in video & 32 & low & -- \\
    Gemini 3.1 Pro & \texttt{gemini-3.1-pro-preview} & API & in video & 32 & low & -- \\
    \bottomrule
  \end{tabular}}
\end{table*}

\section{Full results}
\label{app:results}
\cref{tab:pergen_t1,tab:pergen_t2,tab:pergen_t3} gives accuracy per generator and clip length with 95\% Wilson intervals.

\cref{tab:t4} breaks the T4 results of \cref{tab:zeroshot} down per part, with the text metrics. In the sweeps (\cref{tab:sweeps}), lowering the learning rate from $10^{-4}$ to $10^{-5}$ raises fully fine-tuned PE-AV's T1 accuracy from $56.4 \pm 13.0$\% to 99.8\%; for LN-tuned PE-AV and HAVIC and LoRA-tuned Qwen2.5-Omni, T1 accuracy improves up to the protocol learning rate and varies by at most two points with 1.5--2$\times$ the epoch budget, while halving the budget costs Qwen2.5-Omni 15 points.

\cref{tab:length} gives the results of \cref{tab:zeroshot} split by clip length (3\,s vs.\ 6\,s). \Cref{tab:length_ctrl} repeats the T2 comparison with Wan2.2-Animate removed from the 3\,s clips.

\begin{table*}[t]
  \centering
  \scriptsize
  \renewcommand{\arraystretch}{0.85}
  \caption{Overall results by clip length, 3\,s (77 frames) vs.\ 6\,s (153 frames), on the \mozaic{} and \mozaicsmall{} test sets. T1--T3: accuracy (\%); T3 is the 4-way Joint answer. T4 (on \mozaicsmall{} only): judge score (0--10). Best model per column in \textbf{bold}, second \underline{underlined}.}
  \label{tab:length}
  \resizebox{\linewidth}{!}{%
  \begin{tabular}{ll | cccccc | cccccccc}
    \toprule
    & & \multicolumn{6}{c}{\mozaic{}} & \multicolumn{8}{c}{\mozaicsmall{}} \\
    \cmidrule(lr){3-8} \cmidrule(lr){9-16}
    & & \multicolumn{2}{c}{T1} & \multicolumn{2}{c}{T2} & \multicolumn{2}{c}{T3}
      & \multicolumn{2}{c}{T1} & \multicolumn{2}{c}{T2} & \multicolumn{2}{c}{T3} & \multicolumn{2}{c}{T4} \\
    \cmidrule(lr){3-4} \cmidrule(lr){5-6} \cmidrule(lr){7-8}
    \cmidrule(lr){9-10} \cmidrule(lr){11-12} \cmidrule(lr){13-14} \cmidrule(lr){15-16}
    Model & Size & 3s & 6s & 3s & 6s & 3s & 6s & 3s & 6s & 3s & 6s & 3s & 6s & 3s & 6s \\
    \midrule
    \multicolumn{16}{l}{\textit{Open models}} \\
    Qwen3-Omni           & 30B-A3B & 17.3 & 20.0 & 31.2 & 37.7 & 26.8 & 29.9 & 17.3 & 20.9 & 30.7 & 37.6 & 26.7 & 30.1 & 4.11 & 4.36 \\
    Qwen2.5-Omni         & 7B     & 19.2 & 20.1 & 31.2 & 36.6 & 32.9 & 31.4 & 19.3 & 19.8 & 29.9 & 35.2 & 33.2 & 30.9 & 4.29 & 4.59 \\
    MiniCPM-o 4.5        & 9B      & 23.9 & 24.1 & 30.4 & 37.3 & 33.5 & 30.9 & 23.8 & 23.5 & 30.4 & 36.2 & 33.8 & 31.6 & 4.26 & 4.54 \\
    Nemotron 3 Nano Omni & 30B-A3B & 18.1 & 19.4 & 31.1 & 37.5 & 32.6 & 33.3 & 19.2 & 21.4 & 31.0 & 36.8 & 32.8 & 32.4 & 4.38 & 4.64 \\
    Gemma 4              & 12B     & 16.3 & 20.6 & 30.5 & 36.8 & 27.4 & 32.3 & 16.1 & 20.9 & 30.3 & 37.0 & 27.2 & 33.3 & 4.07 & 4.18 \\
    VideoLLaMA2.1-AV     & 7B      & 32.0 & 29.4 & 27.1 & 32.8 & 29.7 & 28.6 & 32.3 & 30.6 & 27.1 & 33.1 & 29.8 & 28.4 & 4.36 & 4.46 \\
    OmniVinci            & 9B      & 15.4 & 18.1 & 27.5 & 33.2 & 31.1 & 28.8 & 15.3 & 18.2 & 27.6 & 33.1 & 31.5 & 28.3 & 4.05 & 4.31 \\
    Uni-MoE-2.0-Omni     & 33B     & 31.2 & 28.1 & 28.8 & 34.2 & 25.4 & 26.9 & 30.4 & 28.2 & 29.1 & 32.1 & 25.6 & 27.3 & 4.23 & 4.47 \\
    Ola                  & 7B      & \underline{34.1} & 30.5 & 30.4 & 35.5 & 21.3 & 23.8 & \underline{34.3} & 30.7 & 30.6 & 34.4 & 20.0 & 24.8 & 3.77 & 4.34 \\
    VITA-1.5             & 7B      & 16.0 & 19.0 & 28.5 & 36.3 & 21.6 & 23.6 & 15.9 & 19.2 & 29.5 & 35.6 & 20.9 & 23.9 & 4.40 & 4.67 \\
    \midrule
    \multicolumn{16}{l}{\textit{Proprietary models}} \\
    Gemini 3.8 Flash     & --      & 30.3 & \underline{33.1} & \textbf{37.4} & \textbf{45.5} & \underline{38.6} & \underline{43.6} & 30.8 & \underline{33.6} & \textbf{39.3} & \textbf{44.1} & \underline{38.3} & \underline{43.1} & \textbf{5.06} & \underline{5.13} \\
    Gemini 3.1 Pro       & --      & \textbf{34.5} & \textbf{37.0} & \underline{33.4} & \underline{40.8} & \textbf{46.9} & \textbf{48.3} & \textbf{35.0} & \textbf{36.8} & \underline{34.1} & \underline{40.6} & \textbf{46.5} & \textbf{47.8} & \underline{4.97} & \textbf{5.37} \\
    \bottomrule
  \end{tabular}}
\end{table*}

\begin{table}[t]
  \centering
  \scriptsize
  \caption{Clip length under matched generator composition. T2 accuracy (\%) on the \mozaic{} test set for 3\,s clips, 3\,s clips without Wan2.2-Animate (which has no 6\,s clips), and 6\,s clips. $\Delta$: 6\,s minus 3\,s without Animate, computed before rounding. The naive 3\,s to 6\,s gain (\cref{tab:length}) is 5--8 points for every model.}
  \label{tab:length_ctrl}
  \setlength{\tabcolsep}{5pt}
    \resizebox{\linewidth}{!}{%
  \begin{tabular}{lcccc}
    \toprule
    Model & 3\,s & 3\,s w/o Animate & 6\,s & $\Delta$ \\
    \midrule
    Qwen3-Omni           & 31.2 & 37.7 & 37.7 & $+0.1$ \\
    Qwen2.5-Omni         & 31.2 & 37.6 & 36.6 & $-1.1$ \\
    MiniCPM-o 4.5        & 30.4 & 36.7 & 37.3 & $+0.6$ \\
    Nemotron 3 Nano Omni & 31.1 & 37.5 & 37.5 & $0.0$ \\
    Gemma 4              & 30.5 & 36.7 & 36.8 & $0.0$ \\
    VideoLLaMA2.1-AV     & 27.1 & 32.6 & 32.8 & $+0.2$ \\
    OmniVinci            & 27.5 & 33.2 & 33.2 & $+0.1$ \\
    Uni-MoE-2.0-Omni     & 28.8 & 34.7 & 34.2 & $-0.5$ \\
    Ola                  & 30.4 & 36.7 & 35.5 & $-1.2$ \\
    VITA-1.5             & 28.5 & 34.4 & 36.3 & $+2.0$ \\
    \midrule
    Gemini 3.8 Flash     & 37.4 & 45.0 & 45.5 & $+0.5$ \\
    Gemini 3.1 Pro       & 33.4 & 40.2 & 40.8 & $+0.6$ \\
    \bottomrule
  \end{tabular}
  }
\end{table}

\begin{figure}[t]
  \centering
  \begin{minipage}{0.49\linewidth}\centering
    {\footnotesize Ola}\\[1pt]
    \includegraphics[width=\linewidth]{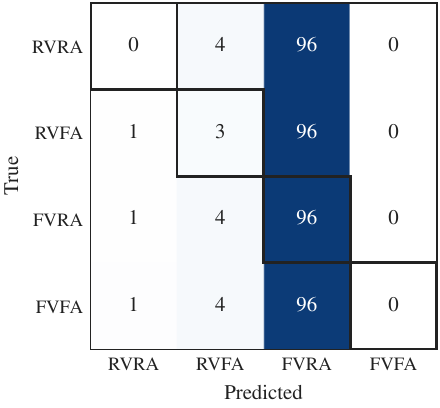}
  \end{minipage}\hfill
  \begin{minipage}{0.49\linewidth}\centering
    {\footnotesize Gemini 3.1 Pro}\\[1pt]
    \includegraphics[width=\linewidth]{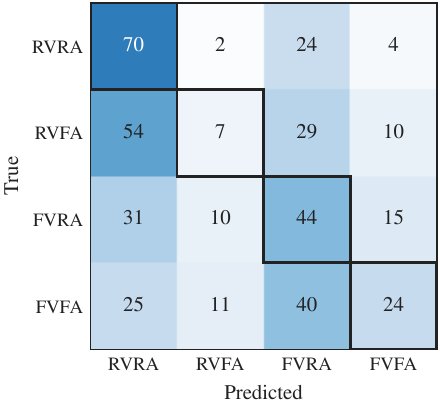}
  \end{minipage}
  \caption{T1 confusion matrices on the \mozaic{} test set: the best open model by T1 accuracy
           (Ola) and the best overall (Gemini 3.1 Pro). Each row is a true label and sums to 100;
           cells give the \% of its clips answered with each label, and the diagonal (outlined) is
           correct. Ola answers FVRA for 96\% of clips whatever the label, so its 32.5\% accuracy comes
           almost entirely from the FVRA clips. Gemini spreads its answers, but most often calls fake audio real
           (RVFA $\to$ RVRA, FVFA $\to$ FVRA).}
  \label{fig:confusion}
\end{figure}

\Cref{tab:audioablation} gives the modality ablation discussed in the main paper, and \cref{fig:confusion} the two confusion matrices it discusses.

\begin{table}[t]
  \centering
  \scriptsize
  \caption{Modality ablation: T1 on the \mozaicsmall{} test set, 32 frames. Acc and F1: 4-way accuracy and macro-F1 (\%). Video, Audio: balanced real/fake accuracy of that modality (\%; chance 50). V only: frames without sound; A only: sound without frames.}
  \label{tab:audioablation}
  \resizebox{\linewidth}{!}{%
  \begin{tabular}{llcccc}
    \toprule
    Model & Input & Acc & F1 & Video & Audio \\
    \midrule
    \multirow{3}{*}{Gemini 3.8 Flash}     & A+V    & 32.0 & 24.8 & 64.0 & 52.1 \\
                                          & V only & 31.1 & 22.0 & 59.4 & 50.1 \\
                                          & A only & 33.2 & 17.4 & 50.2 & 51.8 \\
    \midrule
    \multirow{3}{*}{Nemotron 3 Nano Omni} & A+V    & 20.2 & 17.8 & 51.8 & 48.9 \\
                                          & V only & 23.7 & 16.0 & 53.1 & 50.0 \\
                                          & A only & 16.2 &  9.4 & 49.8 & 49.0 \\
    \midrule
    \multirow{3}{*}{Ola}                  & A+V    & 32.7 & 13.9 & 49.9 & 49.9 \\
                                          & V only & 33.4 & 12.5 & 50.0 & 50.0 \\
                                          & A only & 20.3 & 13.1 & 50.5 & 51.3 \\
    \midrule
    \multirow{3}{*}{MiniCPM-o 4.5}        & A+V    & 23.7 & 18.0 & 52.8 & 50.4 \\
                                          & V only & 20.4 & 13.3 & 52.2 & 50.0 \\
                                          & A only & 21.9 & 18.3 & 49.7 & 45.9 \\
    \midrule
    \multirow{3}{*}{Qwen3-Omni}           & A+V    & 18.9 & 14.4 & 51.5 & 51.8 \\
                                          & V only & 22.4 & 15.2 & 52.5 & 50.0 \\
                                          & A only & 19.1 & 14.0 & 50.0 & 58.6 \\
    \bottomrule
  \end{tabular}}
\end{table}

\cref{tab:confusion_all} gives the T1 confusion matrix of every untuned MLLM, of which \cref{fig:confusion} shows two. Most open models give the same answer for most clips, whatever the label: RVRA (Qwen3-Omni, Qwen2.5-Omni, Gemma~4, OmniVinci, VITA-1.5) or FVRA (Ola, VideoLLaMA2.1-AV, Uni-MoE-2.0-Omni). The latter three therefore have the highest T1 accuracy of the open models: FVRA is a third of the clips.

\begin{table*}[t]
  \centering
  \scriptsize
  \caption{T1 confusion matrices of every untuned MLLM on the \mozaic{} test set (\%). Each row is a true label and sums to 100 (up to rounding); the correct answer (diagonal) is in bold. \Cref{fig:confusion} shows two of them. VideoLLaMA2.1-AV is its nopad run, as in \cref{tab:zeroshot}.}
  \label{tab:confusion_all}
  \setlength{\tabcolsep}{3.5pt}
    \begin{tabular}{llcccc}
      \toprule
      & & \multicolumn{4}{c}{Predicted} \\
      \cmidrule(lr){3-6}
      Model & True & RVRA & RVFA & FVRA & FVFA \\
      \midrule
      \multirow{4}{*}{Qwen3-Omni} & RVRA & \textbf{85.4} & 11.3 & 3.3 & 0.0 \\
       & RVFA & 81.5 & \textbf{14.7} & 3.7 & 0.0 \\
       & FVRA & 82.4 & 11.9 & \textbf{5.7} & 0.0 \\
       & FVFA & 77.1 & 16.3 & 6.6 & \textbf{0.0} \\
      \midrule
      \multirow{4}{*}{Qwen2.5-Omni} & RVRA & \textbf{89.5} & 2.0 & 8.5 & 0.0 \\
       & RVFA & 90.8 & \textbf{2.8} & 6.4 & 0.0 \\
       & FVRA & 84.7 & 2.5 & \textbf{12.9} & 0.0 \\
       & FVFA & 86.1 & 3.2 & 10.7 & \textbf{0.0} \\
      \midrule
      \multirow{4}{*}{MiniCPM-o 4.5} & RVRA & \textbf{64.2} & 4.4 & 31.4 & 0.0 \\
       & RVFA & 64.9 & \textbf{4.9} & 30.3 & 0.0 \\
       & FVRA & 59.0 & 3.5 & \textbf{37.5} & 0.0 \\
       & FVFA & 57.4 & 4.4 & 38.1 & \textbf{0.0} \\
      \midrule
      \multirow{4}{*}{Nemotron 3 Nano Omni} & RVRA & \textbf{56.8} & 34.4 & 8.8 & 0.0 \\
       & RVFA & 59.5 & \textbf{30.1} & 10.4 & 0.0 \\
       & FVRA & 49.3 & 38.0 & \textbf{12.7} & 0.0 \\
       & FVFA & 53.3 & 33.6 & 13.1 & \textbf{0.0} \\
      \midrule
      \multirow{4}{*}{Gemma 4} & RVRA & \textbf{89.5} & 0.8 & 0.4 & 9.3 \\
       & RVFA & 91.1 & \textbf{0.8} & 0.5 & 7.6 \\
       & FVRA & 87.3 & 1.0 & \textbf{0.8} & 10.9 \\
       & FVFA & 89.6 & 0.7 & 0.9 & \textbf{8.7} \\
      \midrule
      \multirow{4}{*}{VideoLLaMA2.1-AV} & RVRA & \textbf{5.8} & 12.7 & 81.3 & 0.2 \\
       & RVFA & 4.1 & \textbf{14.9} & 80.9 & 0.1 \\
       & FVRA & 5.5 & 12.3 & \textbf{82.1} & 0.1 \\
       & FVFA & 4.2 & 15.8 & 79.9 & \textbf{0.1} \\
      \bottomrule
    \end{tabular}
  \hfill
    \begin{tabular}{llcccc}
      \toprule
      & & \multicolumn{4}{c}{Predicted} \\
      \cmidrule(lr){3-6}
      Model & True & RVRA & RVFA & FVRA & FVFA \\
      \midrule
      \multirow{4}{*}{OmniVinci} & RVRA & \textbf{95.0} & 3.1 & 1.9 & 0.0 \\
       & RVFA & 95.4 & \textbf{2.6} & 2.0 & 0.0 \\
       & FVRA & 96.9 & 1.9 & \textbf{1.2} & 0.0 \\
       & FVFA & 96.4 & 2.4 & 1.2 & \textbf{0.0} \\
      \midrule
      \multirow{4}{*}{Uni-MoE-2.0-Omni} & RVRA & \textbf{9.5} & 16.1 & 66.4 & 8.1 \\
       & RVFA & 9.0 & \textbf{14.2} & 67.9 & 8.9 \\
       & FVRA & 8.9 & 13.4 & \textbf{67.0} & 10.7 \\
       & FVFA & 8.3 & 13.3 & 67.9 & \textbf{10.6} \\
      \midrule
      \multirow{4}{*}{Ola} & RVRA & \textbf{0.2} & 3.6 & 96.2 & 0.0 \\
       & RVFA & 0.6 & \textbf{3.0} & 96.3 & 0.0 \\
       & FVRA & 0.7 & 3.6 & \textbf{95.7} & 0.0 \\
       & FVFA & 0.8 & 3.6 & 95.6 & \textbf{0.0} \\
      \midrule
      \multirow{4}{*}{VITA-1.5} & RVRA & \textbf{93.1} & 0.0 & 6.9 & 0.0 \\
       & RVFA & 94.0 & \textbf{0.0} & 6.0 & 0.0 \\
       & FVRA & 94.3 & 0.0 & \textbf{5.6} & 0.0 \\
       & FVFA & 94.0 & 0.0 & 6.0 & \textbf{0.0} \\
      \midrule
      \multirow{4}{*}{Gemini 3.8 Flash} & RVRA & \textbf{64.7} & 6.3 & 28.7 & 0.2 \\
       & RVFA & 59.5 & \textbf{9.5} & 30.2 & 0.9 \\
       & FVRA & 34.6 & 8.2 & \textbf{56.1} & 1.0 \\
       & FVFA & 32.4 & 10.9 & 55.2 & \textbf{1.5} \\
      \midrule
      \multirow{4}{*}{Gemini 3.1 Pro} & RVRA & \textbf{70.3} & 2.1 & 23.7 & 4.0 \\
       & RVFA & 54.2 & \textbf{7.5} & 28.6 & 9.7 \\
       & FVRA & 31.1 & 10.1 & \textbf{43.6} & 15.2 \\
       & FVFA & 24.5 & 11.4 & 39.8 & \textbf{24.3} \\
      \bottomrule
    \end{tabular}
\end{table*}

\begin{table*}[t]
  \centering
  \scriptsize
  \caption{T4 explanation results on the \mozaicsmall{} test set. Judge: mean score from 0 to 10
           per part. Real/fake acc.: share of parts called correctly as real or fake (judge score
           $\geq 5$). Text metrics are computed over both parts together. Real/fake acc., text metrics and
           Parse (replies whose two parts could be read) in \%. ``Always real'' answers
           both parts with the fixed real-recording sentence.}
  \label{tab:t4}
\resizebox{\textwidth}{!}{%
  \begin{tabular}{llcccccccc}
    \toprule
     & & \multicolumn{2}{c}{Judge (0--10)} & \multicolumn{2}{c}{Real/fake acc.\ (\%)} & & & & \\
    \cmidrule(lr){3-4}\cmidrule(lr){5-6}
    Model & Size & Video & Audio & Video & Audio & ROUGE-L & BLEU-4 & CIDEr-D & Parse \\
    \midrule
    Always real        & --      & 4.02 & 5.50 & 33.2 & 50.0 & 20.0 &  0.7 &  0.0 & -- \\
    \midrule
    Qwen3-Omni         & 30B-A3B & 3.62 & 4.83 & 32.7 & 49.1 & 25.3 & 11.2 & 16.4 &  98.7 \\
    Qwen2.5-Omni       & 7B     & 3.72 & 5.13 & 33.3 & 49.9 & 25.1 & 13.5 & 10.1 &  99.9 \\
    MiniCPM-o 4.5      & 9B      & 3.53 & 5.24 & 27.7 & 49.1 & 22.5 &  7.2 &  4.2 & 100.0 \\
    Nemotron 3 Nano Omni & 30B-A3B & 3.95 & 5.05 & 37.4 & 49.7 & 26.5 &  9.9 & 11.3 &  99.9 \\
    Gemma 4            & 12B     & 3.60 & 4.64 & 33.8 & 49.0 & 24.9 & 14.4 & 12.1 &  99.9 \\
    VideoLLaMA2.1-AV   & 7B      & 3.84 & 4.98 & 38.7 & 47.6 & 25.8 &  7.7 &  0.8 &  99.9 \\
    OmniVinci          & 9B      & 3.65 & 4.68 & 33.1 & 42.5 & 23.5 &  8.9 &  4.1 &  99.3 \\
    Uni-MoE-2.0-Omni   & 33B     & 3.63 & 5.04 & 33.5 & 49.1 & 23.5 &  9.2 &  4.3 &  99.3 \\
    Ola                & 7B      & 4.03 & 4.01 & 38.2 & 36.9 & 23.2 &  9.0 &  5.0 & 100.0 \\
    VITA-1.5           & 7B      & 4.03 & 5.01 & 38.2 & 46.6 & 24.4 &  8.3 &  3.1 &  99.5 \\
    \midrule
    Gemini 3.8 Flash   & --      & 4.97 & 5.21 & 57.8 & 50.5 & 24.2 & 10.5 &  4.0 & 100.0 \\
    Gemini 3.1 Pro     & --      & 4.91 & 5.39 & 56.4 & 51.9 & 27.8 & 13.3 &  9.7 & 100.0 \\
    \midrule
    Qwen2.5-Omni (LoRA on T4) & 7B & 5.59 & 8.37 & 65.0 & 86.8 & 45.7 & 29.8 & 53.4 & 100.0 \\
    \bottomrule
  \end{tabular}
  }
\end{table*}

\begin{figure}[t]
  \centering
  \includegraphics[width=\linewidth]{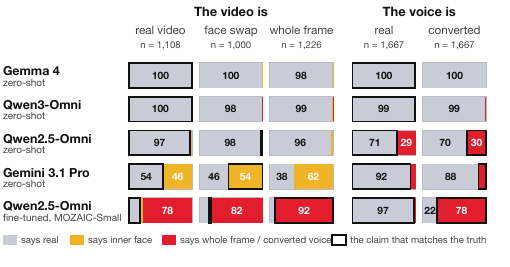}
  \caption{\textbf{What the T4 explanations claim, over the 3,334 \mozaicsmall{} test clips.} Each bar splits a model's answers by the region they assert, for clips grouped by what the video really is (real, inner face swapped, whole frame generated) and what the voice really is (real, converted). The outlined segment is the claim that matches the truth. Open MLLMs say \lq\lq{}real'' for nearly every clip. Gemini~3.1 Pro says \lq\lq{}inner face'' for about half of all clips, real ones included, and never \lq\lq{}whole frame''. The fine-tuned model separates real from converted voices, but says \lq\lq{}whole frame'' for most videos whatever their truth. Claims are parsed from the explanation text.}
  \label{fig:t4_claims}
\end{figure}

\begin{figure*}[t]
  \centering
  \includegraphics[width=\textwidth]{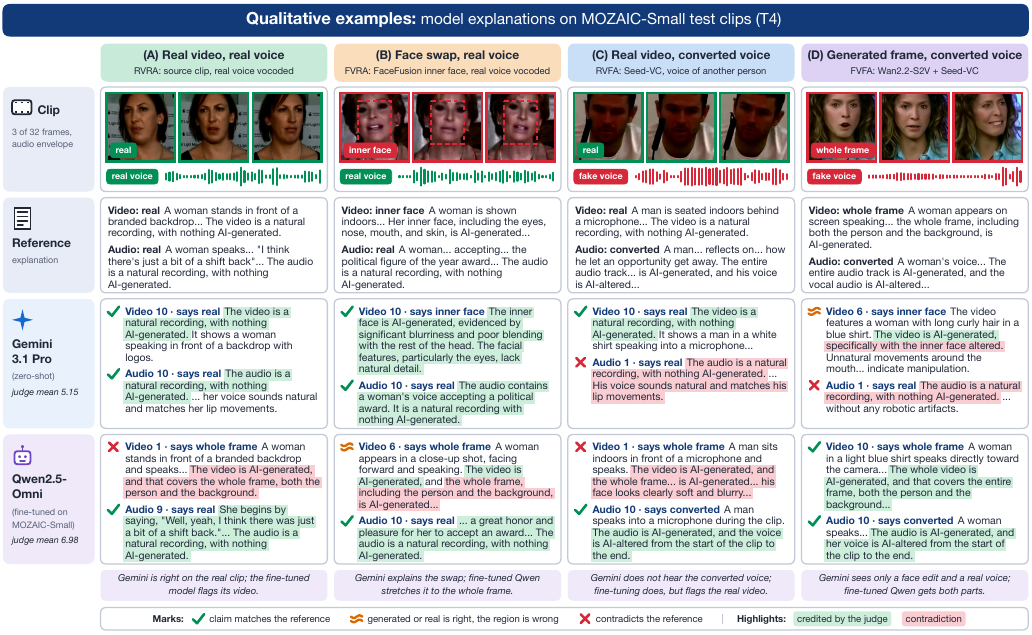}
  \caption{\textbf{T4 explanations in full for the four example clips of the main paper:} real, face swap with real voice, real video with converted voice, and fully generated. Borders and audio envelope: green real, red generated; dashed box: swapped face. Each chip gives the judge's 0--10 score and the claimed region, marked as matching the reference (check), right verdict but wrong region (half check) or contradicting it (cross). Text is a verbatim excerpt, green where credited and red where it contradicts the reference.}
  \label{fig:t4_full}
\end{figure*}

\Cref{fig:t4_full} gives the full explanation texts for the four example clips of the main paper's T4 figure.

\begin{table*}[t]
  \centering
  \scriptsize
  \caption{T1 accuracy (\%) on the \mozaic{} test set by video generator and clip length, with 95\% Wilson intervals. Generator columns hold that generator's fake-video clips; Real video holds the real-video clips.}
  \label{tab:pergen_t1}
  \resizebox{\linewidth}{!}{%
  \begin{tabular}{lccccccc}
    \toprule
    Model & FaceFusion & Wan2.2-S2V & Wan2.2-Animate & Real video & 3\,s & 6\,s & All \\
    \midrule
    Qwen3-Omni & 2.4 (2.0--2.9) & 2.7 (2.2--3.3) & 4.5 (3.5--5.8) & 50.0 (48.6--51.5) & 17.3 (16.5--18.2) & 20.0 (19.0--21.1) & 18.5 (17.9--19.2) \\
    Qwen2.5-Omni & 4.5 (3.9--5.2) & 7.3 (6.5--8.2) & 10.0 (8.5--11.8) & 46.1 (44.7--47.6) & 19.2 (18.4--20.1) & 20.1 (19.1--21.1) & 19.6 (19.0--20.3) \\
    MiniCPM-o 4.5 & 14.9 (13.9--16.1) & 22.2 (20.9--23.6) & 20.9 (18.8--23.2) & 34.5 (33.2--35.9) & 23.9 (23.0--24.9) & 24.1 (23.0--25.2) & 24.0 (23.3--24.7) \\
    Nemotron 3 Nano Omni & 6.3 (5.6--7.1) & 5.8 (5.1--6.6) & 8.2 (6.8--9.8) & 43.5 (42.0--44.9) & 18.1 (17.2--19.0) & 19.4 (18.4--20.5) & 18.7 (18.1--19.4) \\
    Gemma 4 & 2.7 (2.3--3.3) & 8.1 (7.2--9.0) & 1.6 (1.1--2.5) & 45.1 (43.7--46.6) & 16.3 (15.4--17.1) & 20.6 (19.6--21.6) & 18.2 (17.5--18.8) \\
    VideoLLaMA2.1-AV & 41.3 (39.8--42.8) & 41.3 (39.7--42.9) & 40.0 (37.3--42.7) & 10.4 (9.5--11.3) & 32.0 (31.0--33.1) & 29.4 (28.3--30.6) & 30.9 (30.1--31.7) \\
    OmniVinci & 0.7 (0.5--1.0) & 0.5 (0.3--0.7) & 0.8 (0.4--1.4) & 48.8 (47.4--50.3) & 15.4 (14.6--16.2) & 18.1 (17.2--19.1) & 16.6 (16.0--17.3) \\
    Uni-MoE-2.0-Omni & 38.3 (36.9--39.8) & 38.0 (36.4--39.5) & 42.5 (39.8--45.2) & 11.8 (10.9--12.8) & 31.2 (30.1--32.2) & 28.1 (27.0--29.3) & 29.8 (29.0--30.6) \\
    Ola & 47.6 (46.0--49.1) & 48.0 (46.4--49.6) & 48.3 (45.6--51.0) & 1.6 (1.3--2.0) & 34.1 (33.0--35.2) & 30.5 (29.4--31.7) & 32.5 (31.7--33.3) \\
    VITA-1.5 & 3.1 (2.6--3.7) & 2.4 (1.9--2.9) & 3.1 (2.3--4.2) & 46.5 (45.1--48.0) & 16.0 (15.2--16.8) & 19.0 (18.1--20.1) & 17.3 (16.7--18.0) \\
    \midrule
    Gemini 3.8 Flash & 24.1 (22.8--25.4) & 34.9 (33.4--36.5) & 26.0 (23.6--28.4) & 37.1 (35.7--38.5) & 30.3 (29.2--31.3) & 33.1 (32.0--34.4) & 31.6 (30.8--32.4) \\
    Gemini 3.1 Pro & 30.6 (29.2--32.0) & 38.7 (37.1--40.3) & 30.8 (28.4--33.4) & 38.9 (37.5--40.3) & 34.5 (33.4--35.6) & 37.0 (35.8--38.2) & 35.6 (34.8--36.4) \\
    \bottomrule
  \end{tabular}}
\end{table*}

\begin{table*}[t]
  \centering
  \scriptsize
  \caption{T2 accuracy (\%) on the \mozaic{} test set by video generator and clip length, with 95\% Wilson intervals. VideoLLaMA2.1-AV uses its standard run, as T2 has no sound.}
  \label{tab:pergen_t2}
  \resizebox{\linewidth}{!}{%
  \begin{tabular}{lccccccc}
    \toprule
    Model & FaceFusion & Wan2.2-S2V & Wan2.2-Animate & Real video & 3\,s & 6\,s & All \\
    \midrule
    Qwen3-Omni & 40.3 (38.2--42.4) & 0.1 (0.0--0.3) & 0.0 (0.0--0.6) & 66.3 (64.3--68.2) & 31.2 (29.8--32.8) & 37.7 (36.0--39.5) & 34.1 (33.0--35.3) \\
    Qwen2.5-Omni & 13.1 (11.7--14.6) & 0.7 (0.4--1.1) & 0.0 (0.0--0.6) & 88.8 (87.4--90.0) & 31.2 (29.8--32.7) & 36.6 (34.9--38.3) & 33.6 (32.5--34.7) \\
    MiniCPM-o 4.5 & 7.1 (6.0--8.3) & 0.0 (0.0--0.2) & 0.0 (0.0--0.6) & 94.4 (93.3--95.3) & 30.4 (29.0--31.9) & 37.3 (35.6--39.0) & 33.5 (32.4--34.6) \\
    Nemotron 3 Nano Omni & 6.1 (5.2--7.3) & 0.0 (0.0--0.2) & 0.0 (0.0--0.6) & 96.6 (95.8--97.3) & 31.1 (29.6--32.6) & 37.5 (35.8--39.2) & 33.9 (32.8--35.1) \\
    Gemma 4 & 0.4 (0.2--0.8) & 0.1 (0.0--0.4) & 0.0 (0.0--0.6) & 99.7 (99.4--99.9) & 30.5 (29.0--32.0) & 36.8 (35.1--38.5) & 33.3 (32.1--34.4) \\
    VideoLLaMA2.1-AV & 98.2 (97.5--98.7) & 0.4 (0.2--0.9) & 0.3 (0.1--1.1) & 0.0 (0.0--0.2) & 27.1 (25.7--28.5) & 32.8 (31.1--34.5) & 29.6 (28.5--30.7) \\
    OmniVinci & 99.7 (99.3--99.8) & 0.0 (0.0--0.2) & 0.0 (0.0--0.6) & 0.5 (0.3--0.9) & 27.5 (26.1--29.0) & 33.2 (31.6--34.9) & 30.0 (29.0--31.2) \\
    Uni-MoE-2.0-Omni & 76.7 (74.8--78.5) & 1.2 (0.8--1.9) & 0.0 (0.0--0.6) & 23.6 (21.9--25.4) & 28.8 (27.4--30.3) & 34.2 (32.5--35.9) & 31.2 (30.1--32.3) \\
    Ola & 26.9 (25.0--28.9) & 0.0 (0.0--0.2) & 0.0 (0.0--0.6) & 74.2 (72.3--75.9) & 30.4 (29.0--31.9) & 35.5 (33.8--37.3) & 32.7 (31.6--33.8) \\
    VITA-1.5 & 59.1 (57.0--61.2) & 0.0 (0.0--0.2) & 0.0 (0.0--0.6) & 42.9 (40.9--45.0) & 28.5 (27.1--30.0) & 36.3 (34.6--38.1) & 32.0 (30.9--33.1) \\
    \midrule
    Gemini 3.8 Flash & 65.5 (63.4--67.5) & 0.0 (0.0--0.2) & 0.3 (0.1--1.1) & 64.1 (62.1--66.1) & 37.4 (35.8--38.9) & 45.5 (43.7--47.3) & 41.0 (39.8--42.2) \\
    Gemini 3.1 Pro & 89.7 (88.3--90.9) & 0.1 (0.0--0.3) & 0.0 (0.0--0.6) & 29.4 (27.6--31.3) & 33.4 (31.9--34.9) & 40.8 (39.1--42.6) & 36.7 (35.5--37.8) \\
    \bottomrule
  \end{tabular}}
\end{table*}

\begin{table}[t]
  \centering
  \scriptsize
  \caption{T3 joint accuracy (\%) on the \mozaic{} test set by clip length, with 95\% Wilson intervals.}
  \label{tab:pergen_t3}
  \resizebox{\linewidth}{!}{%
  \begin{tabular}{lccc}
    \toprule
    Model & 3\,s & 6\,s & All \\
    \midrule
    Qwen3-Omni & 26.8 (26.0--27.6) & 29.9 (29.0--30.8) & 28.2 (27.6--28.8) \\
    Qwen2.5-Omni & 32.9 (32.0--33.7) & 31.4 (30.6--32.3) & 32.2 (31.6--32.8) \\
    MiniCPM-o 4.5 & 33.5 (32.6--34.3) & 30.9 (30.0--31.8) & 32.3 (31.6--32.9) \\
    Nemotron 3 Nano Omni & 32.6 (31.8--33.4) & 33.3 (32.4--34.2) & 32.9 (32.3--33.5) \\
    Gemma 4 & 27.4 (26.6--28.2) & 32.3 (31.4--33.2) & 29.6 (29.0--30.2) \\
    VideoLLaMA2.1-AV & 29.7 (28.9--30.6) & 28.6 (27.7--29.4) & 29.2 (28.6--29.8) \\
    OmniVinci & 31.1 (30.3--31.9) & 28.8 (28.0--29.7) & 30.0 (29.4--30.6) \\
    Uni-MoE-2.0-Omni & 25.4 (24.6--26.2) & 26.9 (26.1--27.8) & 26.1 (25.5--26.7) \\
    Ola & 21.3 (20.6--22.0) & 23.8 (23.0--24.6) & 22.4 (21.9--23.0) \\
    VITA-1.5 & 21.6 (20.8--22.3) & 23.6 (22.8--24.4) & 22.5 (22.0--23.1) \\
    \midrule
    Gemini 3.8 Flash & 38.6 (37.8--39.5) & 43.6 (42.6--44.5) & 40.9 (40.3--41.6) \\
    Gemini 3.1 Pro & 46.9 (46.0--47.8) & 48.3 (47.3--49.2) & 47.5 (46.9--48.2) \\
    \bottomrule
  \end{tabular}}
\end{table}


\section{Training and compute}
\label{app:compute}
Classifier heads (AV-HuBERT, PE-AV, HAVIC, DFD-HR) are MLPs with dropout 0.1, trained for 10 epochs with learning rate $10^{-4}$, weight decay 0.01, batch size 32, bf16 mixed precision, balanced class weights, single-seed for single runs (five seeds for \cref{tab:seed_stability}), and checkpoints selected on macro-F1. All use 32 frames, with sound at 16\,kHz except PE-AV (\texttt{pe-av-large}, 48\,kHz). DFD-HR keeps CLIP frozen and trains its adapters. Qwen2.5-Omni is fine-tuned with LoRA (rank 16, $\alpha=32$, dropout 0.05) on every linear layer of the language model, with the vision and audio towers frozen, for 3 epochs with learning rate $10^{-4}$, weight decay 0.01 and batch size 32, using a cutoff length of 2,048 tokens on T1 and T2 and 4,096 on T3 and T4. The six other MLLMs tuned for T4 use the same LoRA settings.

\paragraph{Hardware and cost.} Generation, inference and training ran on a shared cluster of NVIDIA A100 and A40 GPUs, one GPU per inference job (two for Qwen3-Omni and Nemotron 3 Nano Omni) and four per classifier training run; the GPU and inference engine used for each model are listed in \cref{tab:models}. The final runs total about 8,800 GPU-hours in the cluster accounting (4,300 on A40, 4,500 on A100), a lower bound since generically named and interactive runs are excluded. Proprietary inference cost about 1,000 USD for the Gemini runs, about 250 USD for the T4 judge, and about 200 USD for the Claude annotation passes.

\begin{table}[t]
  \centering
  \scriptsize
  \caption{\textbf{Seed stability of the trained models.} Mean and standard deviation (subscript) over 5 training seeds at the protocol settings, trained on \mozaicsmall{}. Acc and F1: accuracy and macro-F1 (\%); T3 on the 4-way joint answer. T4: judge score (0--10; overall, video and audio part) and ROUGE-L (\%). Each task lists the models trained on it; on T2, which is video-only, AV-HuBERT and PE-AV use only their visual branch. LN: LayerNorm tuning; Full FT: full fine-tuning.}
  \label{tab:seed_stability}
  \setlength{\tabcolsep}{3.5pt}
  \renewcommand{\arraystretch}{1.1}
  \resizebox{\linewidth}{!}{%
  \begin{tabular}{llcccc}
    \toprule
    & & \multicolumn{2}{c}{\mozaic{}} & \multicolumn{2}{c}{\mozaicsmall{}} \\
    \cmidrule(lr){3-4} \cmidrule(lr){5-6}
    Model & Setting & Acc & F1 & Acc & F1 \\
    \midrule
    \multicolumn{6}{l}{\textit{T1: Audio-Video Authenticity}} \\
    \midrule
    AV-HuBERT     & Full FT  & 58.8\pmstd{7.7} & 43.6\pmstd{5.9} & 59.3\pmstd{7.4} & 43.8\pmstd{6.0} \\
                  & LN       & 58.2\pmstd{2.2} & 50.0\pmstd{1.0} & 58.2\pmstd{2.1} & 50.0\pmstd{0.9} \\
    PE-AV         & Full FT  & 56.4\pmstd{13.0} & 38.0\pmstd{14.5} & 56.6\pmstd{13.1} & 38.0\pmstd{14.5} \\
                  & LN       & 98.4\pmstd{0.1} & 98.4\pmstd{0.1} & 98.8\pmstd{0.3} & 98.7\pmstd{0.2} \\
    HAVIC         & Full FT  & 66.2\pmstd{0.1} & 39.7\pmstd{0.1} & 66.0\pmstd{0.2} & 39.6\pmstd{0.1} \\
                  & LN       & 86.9\pmstd{0.8} & 85.2\pmstd{0.8} & 86.3\pmstd{0.9} & 84.7\pmstd{1.0} \\
    Qwen2.5-Omni  & LoRA     & 83.4\pmstd{1.2} & 81.7\pmstd{1.3} & 83.1\pmstd{1.3} & 81.2\pmstd{1.5} \\
    \addlinespace[3pt]
    \multicolumn{6}{l}{\textit{T2: Manipulation Type}} \\
    \midrule
    AV-HuBERT     & Full FT  & 59.5\pmstd{3.8} & 58.1\pmstd{4.0} & 60.8\pmstd{3.4} & 59.8\pmstd{3.4} \\
                  & LN       & 44.6\pmstd{2.7} & 43.7\pmstd{2.1} & 44.1\pmstd{2.6} & 43.4\pmstd{2.2} \\
    PE-AV         & Full FT  & 31.9\pmstd{1.3} & 17.3\pmstd{3.6} & 32.2\pmstd{1.0} & 17.6\pmstd{3.5} \\
                  & LN       & 96.8\pmstd{0.3} & 95.8\pmstd{0.4} & 96.7\pmstd{0.3} & 95.9\pmstd{0.4} \\
    DFD-HR        & Adapters & 99.9\pmstd{0.0} & 99.8\pmstd{0.0} & 99.9\pmstd{0.0} & 99.8\pmstd{0.1} \\
    Qwen2.5-Omni  & LoRA     & 73.6\pmstd{4.7} & 69.5\pmstd{3.8} & 73.4\pmstd{4.7} & 69.4\pmstd{3.6} \\
    \addlinespace[3pt]
    \multicolumn{6}{l}{\textit{T3: Identity Verification (joint)}} \\
    \midrule
    AV-HuBERT     & Full FT  & 52.4\pmstd{1.0} & 52.8\pmstd{1.1} & 52.8\pmstd{1.1} & 53.2\pmstd{1.1} \\
                  & LN       & 50.7\pmstd{0.4} & 51.4\pmstd{0.5} & 51.0\pmstd{0.3} & 51.6\pmstd{0.5} \\
    PE-AV         & Full FT  & 41.3\pmstd{1.9} & 37.5\pmstd{3.6} & 41.3\pmstd{1.8} & 37.5\pmstd{3.5} \\
                  & LN       & 64.8\pmstd{0.2} & 66.0\pmstd{0.1} & 65.5\pmstd{0.3} & 66.7\pmstd{0.4} \\
    Qwen2.5-Omni  & LoRA     & 75.8\pmstd{1.4} & 76.4\pmstd{1.3} & 75.9\pmstd{1.7} & 76.4\pmstd{1.6} \\
    \addlinespace[3pt]
    \multicolumn{2}{l}{\textit{T4: Explanation (\mozaicsmall{})}} & Judge & Video & Audio & ROUGE-L \\
    \midrule
    Qwen2.5-Omni  & LoRA     & 7.05\pmstd{0.22} & 5.63\pmstd{0.07} & 8.47\pmstd{0.39} & 46.0\pmstd{0.8} \\
    \bottomrule
  \end{tabular}}
\end{table}

\begin{table*}[t]
  \centering
  \scriptsize
  \caption{\textbf{Learning-rate and epoch sweeps of the trained models}, on the \mozaic{} test set (T4: \mozaicsmall{}). Each run changes one setting from the protocol (learning rate $10^{-4}$; 10 epochs for the classifiers, 3 for Qwen2.5-Omni; columns marked $^{\ast}$) and keeps the others. Epoch budgets are relative to the protocol; for Qwen2.5-Omni they are rounded to the nearest whole epoch (1, 3, 5, 6 epochs). T2 is video-only, so AV-HuBERT and PE-AV use only their visual branch. Full FT: full fine-tuning; LN: LayerNorm tuning. \textbf{Bold}: best per row within each sweep.}
  \label{tab:sweeps}
  \setlength{\tabcolsep}{3.5pt}
  \resizebox{\linewidth}{!}{%
  \begin{tabular}{ll|cccccc|cccc}
    \toprule
    & & \multicolumn{6}{c|}{Learning rate} & \multicolumn{4}{c}{Epoch budget} \\
    \cmidrule(lr){3-8} \cmidrule(lr){9-12}
    Model & Setting & $10^{-6}$ & $3{\times}10^{-6}$ & $10^{-5}$ & $3{\times}10^{-5}$ & $10^{-4\,\ast}$ & $3{\times}10^{-4}$ & $0.5{\times}$ & $1{\times}^{\ast}$ & $1.5{\times}$ & $2{\times}$ \\
    \midrule
    \multicolumn{12}{l}{\textit{T1: Audio-Video Authenticity (accuracy \%)}} \\
    AV-HuBERT + head            & Full FT  & 64.8 & \textbf{66.0} & 61.3 & 56.3 & 65.4 & 64.1 & 50.1 & 65.4 & 48.0 & \textbf{70.1} \\
                                & LN       & 51.9 & 56.0 & 56.7 & 54.2 & \textbf{62.1} & 52.3 & 45.0 & \textbf{62.1} & 57.2 & 57.2 \\
    PE-AV + head                & Full FT  & \textbf{99.8} & \textbf{99.8} & \textbf{99.8} & 61.8 & 65.2 & 33.4 & 33.4 & 65.2 & \textbf{66.5} & 66.4 \\
                                & LN       & 46.2 & 65.8 & 83.4 & 93.7 & 98.5 & \textbf{99.0} & 96.6 & 98.5 & 98.7 & \textbf{98.8} \\
    HAVIC + head                & Full FT  & 89.8 & 94.8 & \textbf{95.5} & 84.1 & 66.2 & 66.1 & 66.0 & \textbf{66.2} & \textbf{66.2} & 64.9 \\
                                & LN       & 52.2 & 63.7 & 67.8 & 76.8 & 85.7 & \textbf{86.4} & 84.0 & 85.7 & \textbf{86.2} & \textbf{86.2} \\
    Qwen2.5-Omni                & LoRA SFT & 32.2 & 44.3 & 61.6 & 72.8 & \textbf{83.8} & 82.7 & 69.1 & 83.8 & 84.5 & \textbf{85.7} \\
    \midrule
    \multicolumn{12}{l}{\textit{T2: Manipulation Type (accuracy \%)}} \\
    AV-HuBERT (visual) + head   & Full FT  & 36.9 & 41.5 & 50.5 & 56.2 & \textbf{62.0} & 45.4 & 55.0 & 62.0 & 65.2 & \textbf{65.3} \\
                                & LN       & 25.6 & 27.1 & 31.4 & 40.6 & 44.5 & \textbf{46.9} & 36.2 & 44.5 & 49.1 & \textbf{49.4} \\
    PE-AV (visual) + head       & Full FT  & 99.4 & 99.8 & \textbf{100.0} & 46.2 & 30.2 & 33.2 & 33.2 & 30.2 & 32.4 & \textbf{36.0} \\
                                & LN       & 60.6 & 75.2 & 88.0 & 93.1 & 96.9 & \textbf{97.9} & 95.7 & 96.9 & 97.2 & \textbf{97.9} \\
    DFD-HR                      & Adapters & 87.1 & 97.7 & 99.7 & 99.2 & 99.9 & \textbf{100.0} & 99.8 & 99.9 & \textbf{100.0} & 99.9 \\
    Qwen2.5-Omni                & LoRA SFT & 34.9 & 40.3 & 57.0 & 72.7 & \textbf{78.3} & \textbf{78.3} & 61.5 & 78.3 & 78.9 & \textbf{82.7} \\
    \midrule
    \multicolumn{12}{l}{\textit{T3: Identity Verification (joint accuracy \%)}} \\
    AV-HuBERT + head            & Full FT  & 48.5 & 52.2 & 53.5 & \textbf{54.1} & 52.1 & 30.0 & 51.9 & 52.1 & 52.8 & \textbf{53.4} \\
                                & LN       & 45.4 & 47.5 & 49.3 & 49.9 & 51.2 & \textbf{52.0} & 50.6 & 51.2 & 51.0 & \textbf{51.3} \\
    PE-AV + head                & Full FT  & 61.6 & \textbf{75.9} & 71.8 & 41.4 & 39.9 & 39.9 & 39.8 & 39.9 & \textbf{40.0} & \textbf{40.0} \\
                                & LN       & 42.7 & 52.1 & 59.9 & 60.6 & 64.6 & \textbf{69.0} & 63.0 & 64.6 & \textbf{66.8} & \textbf{66.8} \\
    Qwen2.5-Omni                & LoRA SFT & 37.5 & 46.9 & 66.2 & 74.4 & 75.6 & \textbf{76.9} & 67.2 & 75.6 & \textbf{77.0} & \textbf{77.0} \\
    \midrule
    \multicolumn{12}{l}{\textit{T4: Explanation (judge score, 0--10; \mozaicsmall{})}} \\
    Qwen2.5-Omni                & LoRA SFT & 4.17 & 4.68 & 5.33 & 6.24 & 6.98 & \textbf{7.54} & 5.20 & 6.98 & 7.40 & \textbf{7.43} \\
    \bottomrule
  \end{tabular}}
\end{table*}

\begin{table*}[t]
  \centering
  \scriptsize
  \caption{\textbf{Generalization to unseen generators}, T1 on the \mozaic{} test set. Seen: trained on all three video generators; Unseen: the column's generator is held out from training. Each column group is evaluated on that generator's fake-video clips together with all real-video clips. Acc and F1: 4-way accuracy and macro-F1 (\%). Full FT: full fine-tuning; LN: LayerNorm tuning. \textbf{Bold}: best; \underline{underline}: second-best, per column.}
  \label{tab:seen_unseen_full}
  \setlength{\tabcolsep}{4pt}
  \resizebox{\linewidth}{!}{%
  \begin{tabular}{ll|cccc|cccc|cccc}
    \toprule
    & & \multicolumn{4}{c|}{FaceFusion held out} & \multicolumn{4}{c|}{Wan2.2-S2V held out} & \multicolumn{4}{c}{Wan2.2-Animate held out} \\
    \cmidrule(lr){3-6} \cmidrule(lr){7-10} \cmidrule(lr){11-14}
    & & \multicolumn{2}{c}{Acc} & \multicolumn{2}{c|}{F1} & \multicolumn{2}{c}{Acc} & \multicolumn{2}{c|}{F1} & \multicolumn{2}{c}{Acc} & \multicolumn{2}{c}{F1} \\
    \cmidrule(lr){3-4} \cmidrule(lr){5-6} \cmidrule(lr){7-8} \cmidrule(lr){9-10} \cmidrule(lr){11-12} \cmidrule(lr){13-14}
    Model & Setting & Seen & Unseen & Seen & Unseen & Seen & Unseen & Seen & Unseen & Seen & Unseen & Seen & Unseen \\
    \midrule
    AV-HuBERT + head   & Full FT  & 46.6 & 51.2 & 31.7 & 45.4 & 44.3 & 48.3 & 30.6 & 47.2 & 21.7 & 23.3 & 17.9 & 20.6 \\
                       & LN       & 47.7 & 48.8 & 42.1 & \underline{48.8} & 46.3 & 49.5 & 41.4 & 49.3 & 34.5 & 40.1 & 33.8 & 36.9 \\
    \midrule
    PE-AV + head       & Full FT  & 46.4 & 50.2 & 32.0 & 45.8 & 44.2 & 51.1 & 31.0 & 46.2 & 21.8 & 29.6 & 18.2 & 27.5 \\
                       & LN       & \textbf{98.6} & \textbf{78.3} & \textbf{98.6} & \textbf{76.9} & \textbf{98.6} & \textbf{97.8} & \textbf{98.6} & \textbf{97.8} & \textbf{98.5} & \textbf{98.4} & \textbf{98.3} & \textbf{98.2} \\
    \midrule
    HAVIC + head       & Full FT  & 47.0 & 50.1 & 31.9 & 44.6 & 44.7 & 44.7 & 30.8 & 38.5 & 22.0 & 35.9 & 18.0 & 30.7 \\
                       & LN       & \underline{79.5} & 51.9 & \underline{79.5} & 40.1 & \underline{87.5} & 66.0 & \underline{87.5} & 63.7 & \underline{81.4} & 77.5 & \underline{77.7} & 72.4 \\
    \midrule
    Qwen2.5-Omni       & LoRA SFT & 78.9 & \underline{56.4} & 78.9 & 48.3 & 84.9 & \underline{75.3} & 84.9 & \underline{74.2} & 81.2 & \underline{81.0} & \underline{77.7} & \underline{76.5} \\
    \bottomrule
  \end{tabular}}
\end{table*}

\begin{table*}[t]
  \centering
  \scriptsize
  \caption{Training settings. All trained models use the \mozaicsmall{} training split, with checkpoints selected on its validation split, and start from the official released checkpoints; exact model identifiers are listed in the released code. LN: LayerNorm tuning; Full FT: full fine-tuning. HAVIC's LN setting also trains the token-reduction and pooling layers it adds; DFD-HR keeps CLIP frozen and trains its adapters.}
  \label{tab:training}
  \begin{tabular}{@{}lp{0.42\linewidth}p{0.36\linewidth}@{}}
    \toprule
    Setting & Classifiers: AV-HuBERT~\cite{shi2022avhubert}, PE-AV~\cite{vyas2026pushing}, HAVIC~\cite{peng2026havic}, DFD-HR~\cite{sun2026dfdhr} & Qwen2.5-Omni-7B~\cite{qwen25omni} (LoRA) \\
    \midrule
    Tasks & T1: AV-HuBERT, PE-AV, HAVIC; T2: AV-HuBERT and PE-AV (visual branch), DFD-HR & T1--T4 \\
    Initial weights & AV-HuBERT Large, PE-AV Large, HAVIC (self-supervised), CLIP ViT-L/14 (DFD-HR) & Qwen2.5-Omni-7B \\
    Trained parameters & MLP head (dropout 0.1) with the LayerNorm layers (LN) or all layers (Full FT); DFD-HR: adapters & LoRA (rank 16, $\alpha{=}32$, dropout 0.05) on all linear layers of the language model; vision encoder, audio encoder and projector frozen \\
    Optimizer & AdamW, constant learning rate & AdamW, constant learning rate, no warm-up \\
    Learning rate / weight decay & $10^{-4}$ / 0.01 & $10^{-4}$ / 0.01 \\
    Epochs / batch size (effective) & 10 / 32 & 3 / 32 \\
    Loss & Cross-entropy, balanced class weights & Next-token cross-entropy \\
    Precision / seed & bf16 / 0 & bf16 / 0 \\
    Input & 32 frames; 16\,kHz audio (PE-AV: 48\,kHz) & 32 frames at 224$\times$224; 16\,kHz audio \\
    Max sequence length & n/a & 2,048 tokens (T1, T2); 4,096 (T3, T4) \\
    Checkpoint selection & Validation macro-F1 & Validation macro-F1 (T1--T3); ROUGE-L (T4) \\
    Framework & PyTorch & LLaMA-Factory \\
    \bottomrule
  \end{tabular}
\end{table*}

\section{Release Format}
\label{app:schema}
The release stores each clip at \texttt{videos/}\emph{label}\texttt{/}\emph{generator}\texttt{/}\emph{length}\texttt{/}\emph{arm}\texttt{/}, where the generator is \texttt{real} for RVRA and \texttt{seed\_vc} for RVFA. \texttt{metadata.jsonl} holds one row per clip (\cref{tab:schema_clips}), with the same rows as a typed Parquet table. \texttt{pairs.jsonl} holds one row per pair, with the generators that kept it and its quality-check scores. \texttt{splits.json} maps each identity to its split, \texttt{summary.json} records the settings, quality-check limits and all counts, and \texttt{SHA256SUMS} covers every file. Questions are JSON Lines files, one per task and split (\cref{tab:schema_questions}). \mozaicsmall{} ships the same files for its subset.

\begin{table*}[t]
  \centering
  \scriptsize
  \caption{Fields of the clip metadata (\texttt{metadata.jsonl}), one row per clip. Similarities are cosines, from ECAPA-TDNN for voice and ArcFace for face. Face and picture scores are empty for real video.}
  \label{tab:schema_clips}
    \resizebox{\linewidth}{!}{%
  \begin{tabular}{llp{0.56\textwidth}}
    \toprule
    Field & Type & Meaning \\
    \midrule
    \texttt{file\_name} & string & path of the clip inside the dataset folder \\
    \texttt{label} & string & RVRA, RVFA, FVRA or FVFA \\
    \texttt{video\_generator} & string & \texttt{real}, \texttt{facefusion}, \texttt{s2v} or \texttt{animate} \\
    \texttt{audio\_generator} & string & \texttt{none}, \texttt{seed\_vc} or \texttt{bigvgan} \\
    \texttt{length} & string & \texttt{77f} (3.08\,s) or \texttt{153f} (6.12\,s) \\
    \texttt{arm} & string & \texttt{main} or \texttt{vocoded} \\
    \texttt{split} & string & train, val or test \\
    \texttt{face\_id}, \texttt{voice\_id} & string & VoxCeleb2 identity of the face shown and of the voice heard \\
    \texttt{gender} & string & m or f \\
    \texttt{face\_real}, \texttt{voice\_real} & bool & whether the face and the voice are real \\
    \texttt{pair\_id} & string & the pair the clip belongs to \\
    \texttt{source\_id}, \texttt{target\_id} & string & source identity A and target identity B \\
    \texttt{source\_clip} & string & VoxCeleb2 clip the video was made from \\
    \texttt{target\_face\_clip} & string & target clip the face picture came from \\
    \texttt{target\_voice\_clip} & string & target recording used as the voice reference \\
    \texttt{kept\_by} & list of strings & generators that kept the pair \\
    \texttt{voice\_target}, \texttt{voice\_source} & float & converted voice against the target's reference and the source's real voice \\
    \texttt{voice\_vocoded} & float & vocoded real voice against the real voice \\
    \texttt{face\_target}, \texttt{face\_source} & float or empty & fake face against the target's face picture and the source's real face \\
    \texttt{face\_baseline} & float or empty & source's real face against the target's face picture \\
    \texttt{flicker} & float or empty & brightness change between frames \\
    \texttt{melt}, \texttt{sharp} & float or empty & face wobble and face sharpness, relative to the real clip \\
    \texttt{sha256} & string & checksum of the clip file \\
    \bottomrule
  \end{tabular}
  }
\end{table*}

\begin{table*}[t]
  \centering
  \scriptsize
  \caption{Fields of the question files, one JSON Lines file per task and split.}
  \label{tab:schema_questions}
    \resizebox{\linewidth}{!}{%
  \begin{tabular}{llp{0.56\textwidth}}
    \toprule
    Field & Tasks & Meaning \\
    \midrule
    \texttt{id} & all & unique question id \\
    \texttt{task} & all & task name \\
    \texttt{split} & all & train, val or test \\
    \texttt{clips} & all & clip paths; T3 lists the query clip, then the real reference clip \\
    \texttt{question} & all & question text \\
    \texttt{options} & all & answer options, shuffled for T1--T3; for T4, the region names \\
    \texttt{part} & T3 & face or voice \\
    \texttt{gt\_answer} & all & correct option; for T4, the reference explanation \\
    \texttt{gt\_label} & all & four-way label of the clip (for T3, of the query clip) \\
    \texttt{gt\_target} & all & structured label: fake-video and fake-audio flags (T1), class index (T2), fake and same-person flags (T3), fake parts and regions (T4) \\
    \texttt{gt\_target\_name} & T2, T3 & class name of the answer \\
    \texttt{meta} & all & generators, frame count, arm, source and target identities, pair id \\
    \bottomrule
  \end{tabular}
  }
\end{table*}

\section{Datasheet}
\label{app:datasheet}
\paragraph{Motivation.} \mozaic{} supports training and testing audio-visual deepfake detectors in a controlled way: the same real clip appears in all four classes, and every clip receives the same final encoding.

\paragraph{Composition.} 137,745 clips with sound from 1,000 VoxCeleb2 development-set identities. VoxCeleb2 lists one person under two ids (id04048 and id05211, both in training), so these are 999 different people. Metadata covers each clip's label, generators, length, arm, split, face and voice identity, and gender, and each pair's source and target clips, retained, and generators. Gender comes from VoxCeleb2's metadata; the dataset holds no names.

\paragraph{Collection and processing.} See \cref{sec:dataset} and \cref{app:generation}. Files carry SHA256 checksums, and encoder version text was removed by rewriting the container, without re-encoding.

\paragraph{Limitations.} Clips are short, low-resolution talking-head recordings from YouTube interviews, of mixed quality; a few very dark, greyscale, or peak-limited sources could not reach the normalization targets exactly. Generated video compresses differently from camera video, generated faces have less fine detail, and the voice converter leaves a trace in the sound. The datasheet shipped with the dataset reports these measurements. Three quality-check limits are judgment calls (face at least 0.30, melting at most 2.0, sharpness at least 0.5), and the first batch both informed the limits and is released.

\paragraph{Uses and ethics.} Intended for detection research. Not intended for recognizing people, or commercial use. The people in VoxCeleb2 are public figures in public videos who did not consent to this use; access is gated, research-only, and subject to a removal process. The release includes this datasheet for every clip and data file, and versioned question files. The data could also be used to tune fakes until detectors miss them; the generators we use are already public, and we expect the benefit to detection research to outweigh that risk.

\paragraph{Maintenance and access.} The dataset is maintained by the authors.
The code is released under the Apache License~2.0 at
\href{https://github.com/VectorInstitute/mozaic}{\ghlogo\,GitHub}, and the dataset under
CC~BY-NC~4.0 on \href{https://huggingface.co/datasets/vector-institute/Mozaic}{\hflogo\,Hugging Face}.
Users must credit VoxCeleb2 and the generators, and may not use the data to identify
people. Anyone shown in a clip can ask for it to be removed, and removals are applied
to the released benchmark.

\end{document}